\pdfoutput=1

\documentclass[format=acmsmall,nonacm]{acmart}

\usepackage{
    adjustbox,
    algorithm,
    amsfonts,
    amsmath,
    amsthm,
    booktabs,
    etoolbox,
    footnote,
    graphicx,
    hyperref,
    makecell,
    microtype,
    multirow,
    nicefrac,
    pgfplots,
    pifont,
    siunitx,
    soul,
    threeparttable,
    tikz,
    todonotes,
    xurl,
    wrapfig,
    xcolor
}
\usepackage[noend]{algpseudocode}
\usepackage[inline]{enumitem}
\usepackage[T1]{fontenc}    
\usepackage[utf8]{inputenc} 

\usepackage[labelformat=simple]{subcaption}

\usetikzlibrary{arrows,shapes,positioning,decorations.markings,patterns,calc,matrix}
\usepgfplotslibrary{groupplots}

\newcommand\func[2]{#1{\left(#2\right)}}

\newcommand\sgn[1]{\func{\text{sgn}}{#1}}

\newcommand\mean[1]{\func{\mu}{#1}}
\newcommand\sd[1]{\func{\sigma}{#1}}
\newcommand{\norm}[2]{\left\lVert#1\right\rVert_{#2}}

\usepackage{stackengine}
\stackMath
\newcommand\tsup[2][2]{%
 \def\useanchorwidth{T}%
  \ifnum#1>1%
    \stackon[-.5pt]{\tsup[\numexpr#1-1\relax]{#2}}{\scriptscriptstyle\sim}%
  \else%
    \stackon[.5pt]{#2}{\scriptscriptstyle\sim}%
  \fi%
}
	
\newcommand\yup{\ding{52}}
\newcommand\nope{\ding{56}}

\usepackage{todonotes}

\newcommand\jd[1]{\todo[inline, backgroundcolor=pink]{JD: #1}}
\newcommand\ew[1]{\todo[inline, backgroundcolor=green!25]{EW: #1}}
\newcommand\sat[1]{\todo[inline, backgroundcolor=gray!25]{Sat: #1}}
\newcommand\dm[1]{\todo[inline, backgroundcolor=orange!25]{DM: #1}}

\newcommand{\gc}[1]{\textcolor{blue}{GC: #1}}

    \renewcommand\jd[1]{}
    \renewcommand\ew[1]{}
    \renewcommand\sat[1]{}
    \renewcommand\dm[1]{}
    \renewcommand\gc[1]{}

\robustify\bf
\robustify\makecell

\usepackage{dashbox}
\newcommand\dboxed[1]{\dbox{\ensuremath{#1}}}

\soulregister\ref{7}
\soulregister\subref{7}
\soulregister\cite{7}

\acmJournal{TECS}
\acmVolume{0}
\acmNumber{0}
\acmArticle{0}
\acmYear{2023}
\acmMonth{0}
\acmArticleSeq{0}

\begin{document}

\title{Enabling Binary Neural Network Training on the Edge}

\author{Erwei Wang}
\affiliation{%
    \institution{Imperial College London}
    \city{London}
    \streetaddress{Exhibition Road}
    \postcode{SW7 2BX}
    \country{United Kingdom}
}
\email{erwei.wang@amd.com}

\author{James J. Davis}
\affiliation{%
    \institution{Imperial College London}
    \city{London}
    \streetaddress{Exhibition Road}
    \postcode{SW7 2BX}
    \country{United Kingdom}
}
\email{james.davis@imperial.ac.uk}

\author{Daniele Moro}
\affiliation{%
    \institution{Google}
    \city{Mountain View}
    \streetaddress{2015 Stierlin Court}
    \state{CA}
    \postcode{94043}
    \country{United States}
}
\email{danielemoro@google.com}

\author{Piotr Zielinski}
\affiliation{%
    \institution{Google}
    \city{Mountain View}
    \streetaddress{2015 Stierlin Court}
    \state{CA}
    \postcode{94043}
    \country{United States}
}
\email{zielinski@google.com}

\author{Jia Jie Lim}
\affiliation{%
    \institution{iSize}
    \city{London}
    \streetaddress{107 Cheapside}
    \postcode{EC2V 6DN}
    \country{United Kingdom}
}
\email{jj.lim@isize.co}

\author{Claudionor Coelho}
\affiliation{%
    \institution{Advantest}
    \city{San Jose}
    \streetaddress{3061 Zanker Rd}
    \state{CA}
    \postcode{95134}
    \country{United States}
}
\email{claudionor.coelho@alumni.stanford.edu}

\author{Satrajit Chatterjee}
\affiliation{%
    \city{Palo Alto}
    \state{CA}
    \country{United States}
}
\email{satrajit@gmail.com}

\author{Peter Y. K. Cheung}
\affiliation{%
    \institution{Imperial College London}
    \city{London}
    \streetaddress{Exhibition Road}
    \postcode{SW7 2BX}
    \country{United Kingdom}
}
\email{p.cheung@imperial.ac.uk}

\author{George A. Constantinides}
\affiliation{%
    \institution{Imperial College London}
    \city{London}
    \streetaddress{Exhibition Road}
    \postcode{SW7 2BX}
    \country{United Kingdom}
}
\email{g.constantinides@imperial.ac.uk}

\renewcommand{\shortauthors}{E. Wang et al.}

\begin{abstract}
    The ever-growing computational demands of increasingly complex machine learning models frequently necessitate the use of powerful cloud-based infrastructure for their training. 
    Binary neural networks are known to be promising candidates for on-device inference due to their extreme compute and memory savings over higher-precision alternatives.
    However, their existing training methods require the concurrent storage of high-precision activations for all layers, generally making learning on memory-constrained devices infeasible.
    In this article, we demonstrate that the backward propagation operations needed for binary neural network training are strongly robust to quantization, thereby making on-the-edge learning with modern models a practical proposition.
    We introduce a low-cost binary neural network training strategy exhibiting sizable memory footprint reductions while inducing little to no accuracy loss \emph{vs} Courbariaux \& Bengio's standard approach.
    These decreases are primarily enabled through the retention of activations exclusively in binary format.
    Against the latter algorithm, our drop-in replacement sees memory requirement reductions of 3--5$\times$, while reaching similar test accuracy ($\pm 2$~pp) in comparable time, across a range of small-scale models trained to classify popular datasets.
    We also demonstrate from-scratch ImageNet training of binarized ResNet-18, achieving a 3.78$\times$ memory reduction.
    Our work is open-source, and includes the Raspberry Pi-targeted prototype we used to verify our modeled memory decreases and capture the associated energy drops.
    Such savings will allow for unnecessary cloud offloading to be avoided, reducing latency, increasing energy efficiency, and safeguarding end-user privacy.
\end{abstract}



\maketitle

\section{Introduction}

    Although binary neural networks (BNNs) feature weights and activations with just single-bit precision, many models are able to reach accuracy indistinguishable from that of their higher-precision counterparts~\cite{BNN_CNN_BinaryNet,CSUR}.
    Since BNNs are functionally complete, their limited precision does not impose an upper bound on achievable accuracy~\cite{George_RoyalSociety}.
    BNNs represent the ideal class of neural network for edge inference, particularly for custom hardware implementation, due to their use of XNOR for multiplication: a fast and cheap operation to perform.
    Their compact weights also suit systems with limited memory and increases opportunities for caching, providing further potential performance boosts.
    FINN, the seminal BNN implementation for field-programmable gate arrays, reached the highest CIFAR-10, and SVHN classification rates to date at the time of its publication~\cite{BNN_CNN_FINN}.
    
    \begin{table*}
		\centering
		\caption{Applied approximations used in low-cost neural network training works.
		{\nope} signifies approximation-free ({\tt float32}) variables.}
		\begin{threeparttable}
			\begin{tabular}{cccccccc}
				\toprule
														& \multicolumn{2}{c}{Weights}	& \multirow{2}[1]{*}{\makecell{Weight\\gradients}}	& \multicolumn{2}{c}{Activations}	& \multirow{2}[1]{*}{\makecell{Activation\\gradients}}	& \multirow{2}[1]{*}{\makecell{Batch\\norm.}}	\\
				\cmidrule(lr){2-3} \cmidrule(lr){5-6}
				 & Forward & Backward & & Forward & Backward  
				\\
				\midrule
				\cite{BNN_CNN_DoReFa-Net}				& {\tt int6}\tnote{1}   & {\tt int6}		& {\tt int6}									& {\tt int6}	& {\tt int6}				& {\tt int6}										& \nope												\\
				\cite{gradient_checkpointing_google}	& \nope	    & \nope					& \nope											& \nope	& Recomputed\tnote{2}		& \nope												& \nope												\\
				\cite{fp16_nvidia}						& {\tt float16}	    & {\tt float16}			& {\tt float16}									& {\tt float16}		& {\tt float16}			& {\tt float16}										& \nope												\\
				\cite{low_precision_bn_activations}	& \nope	    & \nope					& \nope											& {\tt int}		& {\tt int}				& \nope												& \nope												\\
				\cite{e5m2_format_mac_training}	& \nope	    & \nope					& \nope											& {\tt E5M2}		& {\tt E5M2}				& \nope												& \nope												\\
				\cite{signsgd}							& \nope	    & \nope					& {\tt bool}									& \nope	   	& \nope					& \nope												& \nope												\\
				\cite{l1_bn_wu}						& \nope	    & \nope					& \nope											& \nope		& \nope					& \nope												& $\ell_1$												\\
				\bf \makecell{This \\ work}						& {\tt bool}    & {\tt float16}				& {\tt bool}									& {\tt bool}	& {\tt bool}				& {\tt float16}								& BNN-specific							\\
				\bottomrule
			\end{tabular}
			\begin{tablenotes}
				\footnotesize
				\item[1] Arbitrary precision was supported, but significant accuracy degradation was observed below 6 bits.
				\item[2] Activations were not retained between forward and backward propagation in order to save memory.
			\end{tablenotes}
		\end{threeparttable}
		\label{tab:features}
	\end{table*}
    
    Despite featuring binary forward propagation, existing BNN training approaches perform backward propagation using high-precision floating-point data types---typically {\tt float32}---often making training infeasible on memory-constrained devices.
    The high-precision activations used between forward and backward propagation commonly constitute the largest proportion of the total memory footprint of a training run~\cite{low_memory_training_technical_report,TinyTL_cvpr}.
    Our understanding of standard BNN training algorithms led us to the following realization: high-precision activations should not be used since we are only concerned with weights and activations' {\em signs}.
    In this article, we present a low-memory BNN training scheme based on this intuition featuring binary activations only, facilitated through batch normalization modification.
    
    By increasing the viability of learning on the edge, this work will reduce the domain mismatch between training and inference---particularly in conjunction with federated learning~\cite{McMahan17,google_federated_learning}---and ensure privacy for sensitive applications~\cite{Agarwa18}.
    Via the aggressive memory footprint reductions they facilitate, our proposals will enable models to be trained without the network access reliance, latency and energy overheads or data divulgence inherent to cloud offloading.
    Herein, we make the following contributions.
    \begin{itemize}
	    \item
            We conduct a variable representation and lifetime analysis of Courbariaux \& Bengio's standard BNN training process~\cite{BNN_CNN_BinaryNet}.
            We use this to identify opportunities for memory savings through approximation.
        \item
            Via our proposed BNN-specific forward and backward batch normalization operations, we implement a neural network training regime featuring all-binary activations.
            This significantly reduces the greatest constituent of a given training run's total memory footprint.
        \item
            We present a successful combination of binary activations and binary weight gradients during neural network training.
            This aggregation allows for further reductions in memory footprint.
        \item
            We systematically evaluate the impact of each of our approximations, and provide a detailed characterization of our scheme's memory requirements \emph{vs} accuracy.
        \item
            Against the standard approach, we report memory reductions of up to 5.44$\times$, with little to no accuracy or convergence rate degradation, when training BNNs to classify MNIST, CIFAR-10, and SVHN.
            No hyperparameter tuning is required.
            We also show that the batch size used can be increased by $\sim$10$\times$ while remaining within a given memory envelope, and even demonstrate the efficacy of ImageNet training as a hard target.
        \item
            We provide an open-source release of our Keras-based training software, memory modeling tool, and Raspberry Pi-targeted prototype for the community to use and build upon\footnote{\url{https://github.com/awai54st/Enabling-Binary-Neural-Network-Training-on-the-Edge}}.
            Our memory breakdown analysis represents a clear road map to further, future reductions.
	\end{itemize}
	
\section{Related Work}\label{sec:related_work}

    The authors of all published works on BNN inference acceleration to date made use of high-precision floating-point data types during training~\cite{BNN_CNN_BinaryConnect,BNN_CNN_BinaryNet,BNN_CNN_ABC-Net,BNN_CNN_REBNET_FCCM,birealnet,LUTNET,LUTNET_TC,logicnets,proxybnn,reactnet}.
    There is precedent, however, for the use of quantization when training non-binary networks, as we show in Table~\ref{tab:features} via side-by-side comparison of the approximation approaches taken in those works along with those detailed in this article.
    
    The effects of quantizing the gradients of models with high-precision data, either fixed or floating point, have been studied extensively.
    Zhou \emph{et al.}~\cite{BNN_CNN_DoReFa-Net} and Wu \emph{et al.}~\cite{R_CNN_FXP_WAGE} trained networks with fixed-point weights and activations using fixed-point gradients, reporting no accuracy loss for AlexNet classifying ImageNet with gradients wider than five bits.
    Wen \emph{et al.}~\cite{terngrad} and Bernstein \emph{et al.}~\cite{signsgd} focused solely on aggressive weight gradient quantization, aiming to reduce communication costs for distributed learning.
    Weight gradients were losslessly quantized into ternary and binary formats, respectively, with forward propagation and activation gradients kept at high precision.
    Tatsumi et al. identified redundancy in the rounding implementations of IEEE-754 standard, such as the IEEE-754 conversion for rounding, subnormal, and not-a-number and infinity encodings, at MAC outputs~\cite{e5m2_format_mac_training}.
    The authors also presented empirical evidences showing the feasibility of training DNNs using low-precision floating point formats such as {\tt E5M1} and {\tt E5M2} which use five bits for exponent and one and two bits for mantissa, respectively.
    In this work, we make the novel observation that BNNs are more robust to approximation during training than higher-precision networks.
    We thus propose a data representation scheme more aggressive than all of the aforementioned works combined, delivering large memory savings with near-lossless performance.
    
    An intuitive method to lower the memory footprint of training is to simply reduce the batch size.
    However, doing so generally leads to increased total training time due to reduced memory reuse~\cite{low_memory_training_technical_report}.
    The method we propose in this article does not conflict with batch size tuning, and further allows the use of large batches while remaining within the memory limits of edge devices.

    Gradient checkpointing---the recomputation of activations during backward propagation---has been proposed as a method to reduce the memory consumption of training~\cite{gradient_checkpointing_chen,gradient_checkpointing_google}.
    Such methods introduce additional forward passes, however, and so increase each run's duration and energy cost.
    Graham~\cite{low_precision_bn_activations} and Chakrabarti \& Moseley~\cite{approx_act_for_memory_eff} saved memory during training by buffering activations in low-precision formats, achieving comparable accuracy to all-{\tt float32} baselines.
    Wu \emph{et al.}~\cite{l1_bn_wu} and Hoffer \emph{et al.}~\cite{l1_bn_norm_matters} reported reduced computational costs via $\ell_1$ batch normalization.
    Finally, Helwegen~\emph{et al.}~\cite{BOP} asserted that the use of both trainable weights and momenta is superfluous in BNN optimizers, proposing a weightless BNN-specific optimizer, Bop, able to reach the same level of accuracy as Adam.
    We took inspiration from these works in locating sources of redundancy present in standard BNN training schemes, and propose BNN-specific modifications to $\ell_1$ batch normalization allowing for activation quantization all the way to binary, thus saving memory without increasing latency.
    Yayla \emph{et al.}~\cite{yayla2022memory} further developed methods to compress the momentum values uniquely introduced in Bop, and obtained memory savings in BNN training without incurring significant loss in accuracy.
    Our method aims to identify common bottlenecks for BNN training, irrespective of the optimizer choice, and is therefore orthogonal and complementary to techniques such as Yayla \emph{et al.}'s.
    
    Recent efforts have shown that, in some circumstances, batch normalization can be completely removed from BNN training.
    Chen~\emph{et al.} replaced the trainable scaling factors and biases within standard $\ell_2$ batch normalization with hand-tuned values, thereby approximating these functions via trial and error~\cite{bnn_without_bn}.
    Our method follows a conventional training approach; no manual, offline steps are required.
    Jiang~\emph{et al.} proposed the use of batch normalization-free BNNs for super-resolution imaging~\cite{bnn_without_bn_for_sr}.
    The information loss incurred from the removal of batch normalization in this case is recovered by expanding the receptive fields of convolution operations using parallel sets of binary dilated convolutions.
    While Jiang~\emph{et al.} demonstrated promising results for super-resolution imaging, we assume a generic deep learning setting rather than focusing on a specific application domain.
    We further present an open-source Raspberry Pi-based prototype to corroborate our memory reduction estimates, making our work closer to real application deployment than both of the aforementioned publications.
    
    The authors of works including Bi-Real Net~\cite{birealnet}, ResNetE-18~\cite{back_to_simplicity_resnete_18}, and ReActNet~\cite{reactnet} discovered that the accuracy of BNNs can be significantly increased via the addition of high-precision skip connections.
    Many further enhanced BNN performance via improvements to gradient approximation and weight initialization~\cite{bnn_plus,birealnet,back_to_simplicity_resnete_18,reactnet,real_to_binary}.
    Optimizations such as these are intended to increase accuracy: a goal orthogonal to ours of efficiently deploying BNNs on edge-scale devices.
    Nevertheless, we incorporated all of them into our work in order to reach competitive accuracy.
    
    For works such as ReActNet~\cite{reactnet}, BN-Free~\cite{bnn_without_bn}, BN-Free ISR~\cite{bnn_without_bn_for_sr}, and Real-to-Binary~\cite{real_to_binary}, it was found that knowledge distillation---the employment of a high-precision network as a ``teacher'' running alongside a BNN---can greatly improve the performance of the latter's training.
    This method is, however, outside our scope; the teacher would dominate overall memory requirements and thereby make savings with regards to the BNN insignificant.
    
\section{Standard Training Flow}
\label{sec:bnn_preliminaries}
    
    \begin{figure}
        \centering
        \begin{tikzpicture}[thick, >=stealth, node distance=11mm]

    \tikzstyle {gradient} = [rectangle, draw, text centered, minimum height=6mm, gray]
	\tikzstyle {operator} = [circle, minimum width=9mm, align=center, draw]
	\tikzstyle {arrow} = [->]
	\tikzstyle {grad_arrow} = [->, dashed, gray]
    
	\node (x_l) {$\boldsymbol{X}_l$};
	\node [gradient, below of=x_l, yshift=5mm] (dx_l) {$\partial\boldsymbol{X}_l$};
	\node [operator, right=of x_l, xshift=-6mm] (sgnx) {sgn};
	\node [operator, right=of sgnx, xshift=-3mm] (mul) {$\times$};
	\node [operator, right=of mul, xshift=-3mm] (bn) {BN};
	\node [right=of bn, xshift=-6mm] (x_lplus1) {$\boldsymbol{X}_{l+1}$};
	\node [gradient] (dx_lplus1) at (dx_l -| x_lplus1) {$\partial\boldsymbol{X}_{l+1}$};
	\node [operator, below=of mul, yshift=3mm] (sgnw) {sgn};
	\node [below=of sgnw, yshift=6mm] (w) {$\boldsymbol{W}_{l}$};
    
    \draw [arrow] (x_l.east) -- (sgnx.west);
    \draw [arrow] (sgnx.east) -- (mul.west) node [midway, above, anchor=south] (xhat_l) {$\hat{\boldsymbol{X}}_l$};
    \draw [arrow] (mul.east) -- (bn.west) node [midway, above, anchor=south] (y) {$\boldsymbol{Y}_l$};
    \draw [arrow] (bn.east) -- (x_lplus1.west);
    \draw [arrow] (sgnw.north) -- (mul.south) node [midway, left, anchor=east] {$\hat{\boldsymbol{W}}_l$};
    \draw [arrow] (w.north) -- (sgnw.south);
    
    \node [gradient] (dw) at (w -| xhat_l) {$\partial\boldsymbol{W}_l$};
    \node [gradient] (dy) at (dx_l -| y) {$\partial\boldsymbol{Y}_l$};
    
    \draw [grad_arrow] (dx_lplus1.west) to (dy.east);
    \draw [grad_arrow, red] (bn.south east) to [out=-45, in=-45, looseness=1.5] (dy.south east);
    \draw [grad_arrow] (dy.west) to (dx_l.east);
    \draw [grad_arrow] (sgnw.north west) [out=135, in=-45] to (dx_l.south east);
    \draw [grad_arrow] (dy.south) to [out=-90, in=15] (dw.north east);
    \draw [grad_arrow] (sgnx.south east) to [out=-45, in=90] (dw.north);

\end{tikzpicture}
        \caption{
            Standard BNN training graph for fully connected layer $l$.
            ``sgn'', ``$\times$'', and ``BN'' are sign, matrix multiplication, and batch normalization operations.
            Forward propagation dependencies are shown with solid lines; those for backward passes are dashed.
            High-precision activations must be retained due to the red dependency.
        }
        \label{fig:training_graph}
    \end{figure}
    
    For simplicity of exposition, we assume the use of a multi-layer perceptron (MLP), although the presence of convolutional layers would not change any of the principles that follow.
    We use $\partial$ symbol to represent a gradient with respect to the neural network cost function $C$, such that $\partial x$ denotes gradient $\nicefrac{\partial C}{\partial x}$.
    Let $\boldsymbol{W}_l$ and $\boldsymbol{X}_l$ denote matrices of weights and activations, respectively, in the network's $l$\textsuperscript{th} layer, with $\partial\boldsymbol{W}_l$ and $\partial\boldsymbol{X}_l$ being their gradients.
    For $\boldsymbol{W}_l$, rows and columns span input and output channels, respectively, while for $\boldsymbol{X}_l$ they span a batch's feature maps and their channels.
    Henceforth, we use $\hat{\bullet}$ to denote binary encoding.
    
    Fig.~\ref{fig:training_graph} shows the training graph of a fully connected binary layer.
    A detailed description of the standard BNN training procedure introduced by Courbariaux \& Bengio~\cite{BNN_CNN_BinaryNet} for each batch of $B$ training samples, which we henceforth refer to as a {\em step}, is provided in Algorithm~\ref{alg:bnn_vanilla}.
    Therein, ``$\odot$'' signifies element-wise multiplication.
    For brevity, we omit some of the intricacies of the baseline implementation---lack of first-layer quantization, use of a final softmax layer, and the inclusion of weight gradient cancelation~\cite{BNN_CNN_BinaryNet}---as these standard BNN practices are not impacted by our work.
    We initialize weights as outlined by Glorot \& Bengio~\cite{glorot_initialisation}.
    
    \begin{figure*}
        \vspace{-1em}
        \begin{minipage}[t]{0.49\textwidth}
            \begin{algorithm}[H]
                \caption{Standard BNN training step.}
                \begin{algorithmic}[1]
                    \For{$l \leftarrow \left\{1,\cdots,L-1\right\}$}\Comment{Forward prop.}
                        \State $\hat{\boldsymbol{X}}_l \leftarrow \sgn{\boldsymbol{X}_l}$ \label{alg:bnn_vanilla:sgnx}
                        \State $\hat{\boldsymbol{W}}_l \leftarrow \sgn{\boldsymbol{W}_l}$
                        \State $\boldsymbol{Y}_l \leftarrow \hat{\boldsymbol{X}}_l \hat{\boldsymbol{W}}_l$
                        \For{$m \leftarrow \left\{1,\cdots,M_l\right\}$} \label{alg:bnn_vanilla:bn_fwd_start}\Comment{Batch norm.}
                            \State $\psi^{\left(m\right)}_l \leftarrow \sd{\boldsymbol{y}^{\left(m\right)}_l}$
                            \State $\boldsymbol{x}^{\left(m\right)}_{l+1} \leftarrow \frac{\boldsymbol{y}^{\left(m\right)}_l - \mean{\boldsymbol{y}^{\left(m\right)}_l}}{\psi^{\left(m\right)}_l} + \beta^{\left(m\right)}_l$ \label{alg:bnn_vanilla:bn_fwd_end}
                            \State ${\color{white}\omega^{\left(m\right)}_{l+1} \leftarrow \nicefrac{\norm{\boldsymbol{x}^{\left(m\right)}_{l+1}}{1}}{B}}$
                        \EndFor
                    \EndFor
                    \For{$l \leftarrow \left\{L-1,\cdots,1\right\}$}\Comment{Backward prop.}
                        \For{$m \leftarrow \left\{1,\cdots,M_l\right\}$} \label{alg:bnn_vanilla:bn_bwd_start}\Comment{Batch norm.}
                            \State $\boldsymbol{v} \leftarrow \frac{1}{\psi^{\left(m\right)}_l}\partial\boldsymbol{x}^{\left(m\right)}_{l+1}$ \label{alg:bnn_vanilla:bn_bwd_v}
                            \State $\partial\boldsymbol{y}^{\left(m\right)}_l \leftarrow \boldsymbol{v} - \mean{\boldsymbol{v}} - \mean{\boldsymbol{v} \odot \dboxed{\boldsymbol{x}^{\left(m\right)}_{l+1}}}\dboxed{\boldsymbol{x}^{\left(m\right)}_{l+1}}$ \label{alg:bnn_vanilla:bn_bwd_main}
                            \State $\partial\beta^{\left(m\right)}_l \leftarrow \sum{\partial\boldsymbol{x}^{\left(m\right)}_{l+1}}$ \label{alg:bnn_vanilla:bn_bwd_end}
                        \EndFor
                        \State $\partial\boldsymbol{X}_l \leftarrow \partial\boldsymbol{Y}_l\hat{\boldsymbol{W}}_l^\text{T}$ \label{alg:bnn_vanilla_ste_x}
                        \State $\partial\boldsymbol{W}_l \leftarrow \hat{\boldsymbol{X}}_l^\text{T} \partial\boldsymbol{Y}_l$ \label{alg:bnn_vanilla_ste_w}
                        \State ${\color{white}\partial\hat{\boldsymbol{W}}_l \leftarrow \sgn{\partial\boldsymbol{W}_l}}$
                    \EndFor
                    \For{$l \leftarrow \left\{1,\cdots,L-1\right\}$}\Comment{Weight update}
                        \State $\boldsymbol{W}_l \leftarrow \func{\text{Optimize}}{\boldsymbol{W}_l,\partial\boldsymbol{W}_l,\eta}$
                        \State $\boldsymbol{\beta}_l \leftarrow \func{\text{Optimize}}{\boldsymbol{\beta}_l,\partial\boldsymbol{\beta}_l,\eta}$ \label{alg:bnn_vanilla_lr}
                    \EndFor
                    \State $\eta \leftarrow \func{\text{LearningRateSchedule}}{\eta}$
                \end{algorithmic}
                \label{alg:bnn_vanilla}
            \end{algorithm}
        \end{minipage}%
        \hfill%
        \begin{minipage}[t]{0.49\textwidth}
            \begin{algorithm}[H]
                \caption{Proposed BNN training step.}
                \begin{algorithmic}[1]
                    \For{$l \leftarrow \left\{1,\cdots,L-1\right\}$}\Comment{Forward prop.}
                        \State $\hat{\boldsymbol{X}}_l \leftarrow \sgn{\boldsymbol{X}_l}$
                        \State $\hat{\boldsymbol{W}}_l \leftarrow \sgn{\boldsymbol{W}_l}$
                        \State $\boldsymbol{Y}_l \leftarrow \hat{\boldsymbol{X}}_l \hat{\boldsymbol{W}}_l$
                        \For{$m \leftarrow \left\{1,\cdots,M_l\right\}$} \label{alg:bnn_ours:bn_fwd_start}\Comment{Batch norm.}
                            \vspace{0.1em}
                            \State ${\color{red}\psi^{\left(m\right)}_l \leftarrow \nicefrac{\norm{\boldsymbol{y}^{\left(m\right)}_l - \mean{\boldsymbol{y}^{\left(m\right)}_l}}{1}}{B}}$
                            \vspace{0.35em}
                            \State $\boldsymbol{x}^{\left(m\right)}_{l+1} \leftarrow \frac{\boldsymbol{y}^{\left(m\right)}_l - \mean{\boldsymbol{y}^{\left(m\right)}_l}}{\color{red}\psi^{\left(m\right)}_l} + \beta^{\left(m\right)}_l$ \label{alg:bnn_ours:fwd_x}
                            \State ${\color{red}\omega^{\left(m\right)}_{l+1} \leftarrow \nicefrac{\norm{\boldsymbol{x}^{\left(m\right)}_{l+1}}{1}}{B}}$
                            \label{alg:bnn_ours:bn_fwd_end}
                        \EndFor
                    \EndFor
                    \For{$l \leftarrow \left\{L-1,\cdots,1\right\}$}\Comment{Backward prop.}
                        \For{$m \leftarrow \left\{1,\cdots,M_l\right\}$} \label{alg:bnn_ours:bn_bwd_start}\Comment{Batch norm.}
                            \State $\boldsymbol{v} \leftarrow \frac{1}{\color{red}\psi^{\left(m\right)}_l}\partial\boldsymbol{x}^{\left(m\right)}_{l+1}$
                            \State $\partial\boldsymbol{y}^{\left(m\right)}_l \leftarrow \boldsymbol{v}\!-\!\mean{\boldsymbol{v}}\!-\!\mean{\boldsymbol{v} \odot \dboxed{{\color{red}\hat{\boldsymbol{x}}^{\left(m\right)}_{l+1}\omega^{\left(m\right)}_{l+1}}}}\dboxed{{\color{red}\hat{\boldsymbol{x}}^{\left(m\right)}_{l+1}}}$
                            \State $\partial\beta^{\left(m\right)}_l \leftarrow \sum{\partial\boldsymbol{x}^{\left(m\right)}_{l+1}}$ \label{alg:bnn_ours:bn_bwd_end}
                        \EndFor
                        \State $\partial\boldsymbol{X}_l \leftarrow \partial\boldsymbol{Y}_l\hat{\boldsymbol{W}}_l^\text{T}$ 
                        \label{alg:bnn_ours_ste_x}
                        \State $\partial\boldsymbol{W}_l \leftarrow \hat{\boldsymbol{X}}_l^\text{T} \partial\boldsymbol{Y}_l$ \label{alg:bnn_ours_ste_w}
                        \State ${\color{red}\partial\hat{\boldsymbol{W}}_l \leftarrow \sgn{\partial\boldsymbol{W}_l}}$
                        \label{alg:bnn_ours_bin_dw}
                    \EndFor
                    \For{$l \leftarrow \left\{1,\cdots,L-1\right\}$}\Comment{Weight update}
                        \State $\boldsymbol{W}_l \leftarrow \func{\text{Optimize}}{\boldsymbol{W}_l,{\color{red}\nicefrac{\partial\hat{\boldsymbol{W}}_l} {\sqrt{M_{l-1}}} },\eta}$
                        \label{alg:bnn_ours_opt_w}
                        \State $\boldsymbol{\beta}_l \leftarrow \func{\text{Optimize}}{\boldsymbol{\beta}_l,\partial\boldsymbol{\beta}_l,\eta}$
                    \EndFor
                    \State $\eta \leftarrow \func{\text{LearningRateSchedule}}{\eta}$
                \end{algorithmic}
                \label{alg:bnn_ours}
            \end{algorithm}  
        \end{minipage}
		
		\vspace{0.25em}
		\footnotesize
        $\hat{\bullet}$ denotes binary encoding.
        Our refinements are shown in red.
        Dashed boxes highlight Algorithm~\ref{alg:bnn_ours}'s lack of high-precision activations.
    \end{figure*}
    
    Many authors have established that BNNs require batch normalization in order to avoid gradient explosion~\cite{BNN_OPTIMISER_SURVEY,sari_how_does_bn_help_bnn,BNN_A_SURVEY}, and our early experiments confirmed this to indeed be the case.
    We thus apply it as standard.
    Matrix products $\boldsymbol{Y}_l$ are channel-wise batch-normalized across each layer's $M_l$ output channels (lines~\ref{alg:bnn_vanilla:bn_fwd_start}--\ref{alg:bnn_vanilla:bn_fwd_end}) to form the subsequent layer's inputs, $\boldsymbol{X}_{l+1}$.
    $\boldsymbol{\beta}$ constitutes the batch normalization biases.
    Layer-wise moving means $\mean{\boldsymbol{y}_l}$ and standard deviations $\sd{\boldsymbol{y}_l}$ are retained for use during backward propagation and inference.
    We forgo trainable scaling factors; these are irrelevant to BNNs since their activations are binarized thereafter (line~\ref{alg:bnn_vanilla:sgnx}).
    

    As emphasized in both Fig.~\ref{fig:training_graph} and Algorithm~\ref{alg:bnn_vanilla} (line~\ref{alg:bnn_vanilla:bn_bwd_main}), {\em high-precision} storage of the entire network's activations is required.
    Addressment of this forms our key contribution.
    
\section{Variable Analysis}
    
    In order to quantify the potential gains from approximation, we conducted a variable representation and lifetime analysis of Algorithm~\ref{alg:bnn_vanilla} following the approach taken by Sohoni \emph{et al.}~\cite{low_memory_training_technical_report}.
    Table~\ref{tab:memory_footprint} lists the properties of all variables in Algorithm~\ref{alg:bnn_vanilla}, with each variable's contribution to the total footprint shown for a representative example.
    Variables are divided into two classes: those that must remain in memory between computational phases (forward propagation, backward propagation, and weight update), and those that need not.
    This is of pertinence since, for those in the latter category, only the largest layer's contribution counts towards the total memory occupancy.
    For example, $\partial\boldsymbol{X}_l$ is read during the backward propagation of layer $l-1$ only, thus $\partial\boldsymbol{X}_{l-1}$ can safely overwrite $\partial\boldsymbol{X}_l$ for efficiency.
    Additionally, $\boldsymbol{Y}$ and $\partial\boldsymbol{X}$ are shown together since they are equally sized and only need to reside in memory during the forward and backward pass for each layer, respectively.
    
    \begin{table*}
		\centering
		\caption{Exemplary memory-related properties of variables used during CIFAR-10 training of BinaryNet with Adam and a batch size of 100.}
        \begin{threeparttable}
		    \begin{tabular}{cccSScSS}
				\toprule
				\multirow{3}[2]{*}{Variable} & \multirow{3}[2]{*}{\makecell{Per-layer\\lifetime\tnote{1}}} & \multicolumn{3}{c}{Standard training} & \multicolumn{3}{c}{\bf Proposed training} \\
				\cmidrule(lr){3-5} \cmidrule(lr){6-8}
				 &  & \multirow{2}{*}{\makecell{Data\\type}} & {\multirow{2}{*}{\makecell{Modeled\\memory (MiB)}}} & {\multirow{2}{*}{\%}} & \multirow{2}{*}{\makecell{Data\\type}} & {\multirow{2}{*}{\makecell{Modeled\\memory (MiB)}}} & {\multirow{2}{*}{$\Delta$ ($\times\downarrow$)}} \\
				 &  &  &  &  &  &  &  \\
				\midrule
				$\boldsymbol{X}$ & \nope & {\tt f32} & 111.33 & 21.71 & {\tt bool} & 3.48 & 32.00 \\
				$\partial\boldsymbol{X}$, $\boldsymbol{Y}$\tnote{2} & \yup & {\tt f32} & 50.00 & 9.75 & {\tt f16} & 25.00 & 2.00 \\
				$\mean{\boldsymbol{y}_l}$, $\sd{\boldsymbol{y}_l}$ & \nope &{\tt f32} & 0.03 & 0.00 & {\tt f16} & 0.02 & 2.00 \\
				$\partial\boldsymbol{Y}$ & \yup & {\tt f32} & 50.00 & 9.75 & {\tt f16} & 25.00 & 2.00 \\
				$\boldsymbol{W}$ & \nope & {\tt f32} & 53.49 & 10.43 & {\tt f16} & 26.74 & 2.00 \\
				$\partial\boldsymbol{W}$ & \nope & {\tt f32} & 53.49 & 10.43 & {\tt bool} & 1.67 & 32.00 \\
				$\boldsymbol{\beta}$, $\partial\boldsymbol{\beta}$ & \nope & {\tt f32} & 0.03 & 0.00 & {\tt f16} & 0.02 & 2.00 \\
				Momenta & \nope &{\tt f32} & 106.98 & 20.86 & {\tt f16} & 53.49 & 2.00 \\
				Pooling masks & \nope &{\tt f32} & 87.46 & 17.06 & {\tt bool} & 2.73 & 32.00 \\
				\midrule
				Total &  &  & 512.81 & 100.00 &  & 138.15 & 3.71 \\
				\bottomrule
			\end{tabular}
			\begin{tablenotes}
			    \footnotesize
			    \item[1] \yup~indicates that a variable does not need to be retained between forward, backward or update phases.
			    \item[2] $\partial\boldsymbol{X}$ and $\boldsymbol{Y}$ can share memory since they are equally sized and have non-overlapping lifetimes.
			\end{tablenotes}
		\end{threeparttable}
		\label{tab:memory_footprint}
	\end{table*}

\section{Low-Cost BNN Training}
\label{sec:low-cost_bnn_training}

    As shown in Table~\ref{tab:memory_footprint}, all variables within the standard BNN training flow use {\tt float32} representation.
    In the subsections that follow, we detail the application of aggressive approximation specifically tailored to BNN training.
    Further to this, and in line with the observation by many authors that {\tt float16} can be used for ImageNet training without inducing accuracy loss~\cite{fp16_nvidia,fp8_16_ibm,mixed_precision_training}, we also switch all remaining variables to this format.
    Our final training procedure is captured in Algorithm~\ref{alg:bnn_ours}, with modifications from Algorithm~\ref{alg:bnn_vanilla} in red and the corresponding data representations used shown in Table~\ref{tab:memory_footprint}.
    
    \subsection{Batch Normalization Approximation}
        
        Analysis of the backward pass of Algorithm~\ref{alg:bnn_vanilla} reveals conflicting requirements for the precision of $\boldsymbol{X}$.
        When computing weight gradients $\partial\boldsymbol{W}$ (line~\ref{alg:bnn_vanilla_ste_w}), only binary activations $\hat{\boldsymbol{X}}$ are needed.
        For the batch normalization training (lines~\ref{alg:bnn_vanilla:bn_bwd_start}--\ref{alg:bnn_vanilla:bn_bwd_end}), however, high-precision $\boldsymbol{X}$ is used.
        The latter occurrences are highlighted with dashed boxes.
        Per Table~\ref{tab:memory_footprint}, the storage of $\boldsymbol{X}$ between forward and backward propagation constitutes the single largest portion of the algorithm's total memory.
        If we are able to use $\hat{\boldsymbol{X}}$ in place of $\boldsymbol{X}$ for these operations, there will be no need for this high-precision activation retention, significantly reducing memory footprint as a result.
        We achieve this as follows.
        
        \subsubsection*{Step 1: $\ell_1$ Normalization}
        
            Standard batch normalization sees channel-wise $\ell_2$ normalization performed on each layer's centralized activations.
            Wu~\emph{et al}, however, shows that the less-costly $\ell_1$ normalization is approximately equivalent to the original $\ell_2$ normalization, by proving that $\ell_1$ normalization is approximately equivalent to the original $\ell_2$ normalization multiplied with a fixed scaling factor equal to $\sqrt{\nicefrac{\pi}{2}}$~\cite{l1_bn_wu}.
            We argue that this observation is especially true for BNNs, in which batch normalization is immediately followed by binarization, thus canceling the effects of any scaling factor.
            
			\jd{Aside from the apparent error here (l1 is equivalent to l1?), this is a bit jarring. In the sentence above, we seem to acknowledge the difference between l2 and l1 (``l1 is good enough''), but below the message changes to ``l1 is basically the same as l2''. I suggest picking one of these arguments and only using that. And why does a scaling factor matter, given the above re immediate binarisation?}
			\ew{Agreed. Decided to take the second argument since it is supported by another work. The reviewer was asking for justification of l1.}
            
            Replacement of batch normalization's backward propagation operation with our $\ell_1$ norm-based version sees lines~\ref{alg:bnn_vanilla:bn_bwd_v}--\ref{alg:bnn_vanilla:bn_bwd_main} of Algorithm~\ref{alg:bnn_vanilla} swapped with \eqref{eqn:bn_bwd_core}, where $B$ is the batch size.
            Not only does our use of $\ell_1$ batch normalization transform one occurrence of $\boldsymbol{x}^{\left(m\right)}_{l+1}$ into its binary form, it also beneficially eliminates all squares and square roots.
            
            \begin{equation}
                \begin{split}
                    \boldsymbol{v} &\leftarrow \frac{1}{\nicefrac{\norm{\boldsymbol{y}^{\left(m\right)}_l - \mean{\boldsymbol{y}^{\left(m\right)}_l}}{1}}{B}}\partial\boldsymbol{x}^{\left(m\right)}_{l+1}\\
                    \partial\boldsymbol{y}^{\left(m\right)}_l &\leftarrow \boldsymbol{v} - \mean{\boldsymbol{v}} - \mean{\boldsymbol{v} \odot \boldsymbol{x}^{\left(m\right)}_{l+1}} \hat{\boldsymbol{x}}^{\left(m\right)}_{l+1}
                \end{split}
                \label{eqn:bn_bwd_core}
            \end{equation}
            
            Our derivation of this function is as follows.
            Let
            \begin{equation*}
                \boldsymbol{a} = \boldsymbol{y}_l - \mu{\left({\boldsymbol{y}_l}\right)}
            \end{equation*}
            and    
            \begin{equation*}
                \boldsymbol{v} = \frac{1}{\frac{\left\lVert\boldsymbol{a}\right\rVert}{B}}\frac{\partial C}{\partial\boldsymbol{x}_{l+1}},
            \end{equation*}
            so that our forward function in line~\ref{alg:bnn_ours:fwd_x} becomes
            \begin{equation*}
                \boldsymbol{x}^{\left(m\right)}_{l+1} \leftarrow \frac{a^{\left(m\right)}}{\psi^{\left(m\right)}_l} + \beta^{\left(m\right)}_l.
            \end{equation*}
            We compute the expression for gradient $\nicefrac{\partial C}{\partial\boldsymbol{y}_l}$ by first computing $\nicefrac{\partial C}{\partial\boldsymbol{a}}$, which can be derived with chain rule,
            \begin{align*}
                \frac{\partial C} {\partial a^{\left(m\right)}} &= \frac{\partial C}{\partial\boldsymbol{x}^{\left(m\right)}_{l+1}} \cdot \frac{\partial \boldsymbol{x}^{\left(m\right)}_{l+1}}{\partial a^{\left(m\right)}} + \frac{\partial C}{\partial \psi^{\left(m\right)}_l} \cdot \frac{\partial \psi^{\left(m\right)}_l}{\partial a^{\left(m\right)}} \\
                &= \frac{\partial C}{\partial\boldsymbol{x}^{\left(m\right)}_{l+1}} \cdot \frac{\partial \boldsymbol{x}^{\left(m\right)}_{l+1}}{\partial a^{\left(m\right)}} + \frac{\partial C}{\partial \boldsymbol{x}^{\left(m\right)}_{l+1}} \cdot \frac{\partial \boldsymbol{x}^{\left(m\right)}_{l+1}}{\partial \psi^{\left(m\right)}_l} \cdot \frac{\partial \psi^{\left(m\right)}_l}{\partial a^{\left(m\right)}}.
            \end{align*}
            By evaluating each component in the above equation, we have
            \begin{equation*}
                \frac{\partial C}{\partial\boldsymbol{a}} = \boldsymbol{v} -  \frac{\boldsymbol{a}}{\left|\boldsymbol{a}\right|} \times \frac{1}{\left(\frac{\left\lVert\boldsymbol{a}\right\rVert}{B}\right)^2} \times \mu{\left(\boldsymbol{a} \odot \frac{\partial C}{\partial\boldsymbol{x}_{l+1}}\right)}
            \end{equation*}
            and thus
            \begin{align*}
                \frac{\partial C}{\partial\boldsymbol{y}_l} &= \frac{\partial C}{\partial\boldsymbol{a}} - \mu{\left(\frac{\partial C}{\partial\boldsymbol{a}} \right)} \\
                 &= \left(\boldsymbol{v} - \mu{\left(\boldsymbol{v}\right)}\right) - \\
                 &\qquad \left(\frac{\boldsymbol{x}_{l+1}}{\frac{\left\lVert\boldsymbol{a}\right\rVert}{B} \odot \left|\boldsymbol{a}\right|} - \mu{\left(\frac{\boldsymbol{x}_{l+1}}{\frac{\left\lVert\boldsymbol{a}\right\rVert}{B} \odot \left|\boldsymbol{a}\right|}\right)}\right)\left(\mu{\left(\boldsymbol{a} \odot \frac{\partial C}{\partial\boldsymbol{x}_{l+1}}\right)}\right).
            \end{align*}
            Since the output of batch normalization, $\boldsymbol{x}_{l+1}$, is expected to have a mean value of zero across samples in a batch, \emph{i.e.},
            \begin{equation*}
                \mu{\left(\boldsymbol{x}_{l+1}\right)} \approx \boldsymbol{0},
            \end{equation*}
            we have
            \begin{equation*}
                \frac{\partial C}{\partial\boldsymbol{y}_l} \approx \left(\boldsymbol{v} - \mu{\left(\boldsymbol{v}\right)}\right) - \mu{\left(\boldsymbol{v} \odot \boldsymbol{x}_{l+1}\right)}\hat{\boldsymbol{x}}_{l+1}.
            \end{equation*}

        \subsubsection*{Step 2: BNN-Specific Approximation}
        
            We further replace the remaining $\boldsymbol{x}^{\left(m\right)}_{l+1}$ term in (\ref{eqn:bn_bwd_core}) with the product of its signs and mean magnitude--- $\hat{\boldsymbol{x}}^{\left(m\right)}_{l+1}\omega^{\left(m\right)}_{l+1}$---where $\omega^{\left(m\right)}_{l+1}$ is precomputed (line~\ref{alg:bnn_ours:bn_fwd_end}).
            
            Our complete batch normalization training functions are shown on lines~\ref{alg:bnn_ours:bn_bwd_start}--\ref{alg:bnn_ours:bn_bwd_end} of Algorithm~\ref{alg:bnn_ours}.
            As again highlighted within dashed boxes, these only require the storage of binary $\hat{\boldsymbol{X}}$ along with layer- and channel-wise mean magnitudes.
            With elements of $\boldsymbol{X}$ now binarized, we reduce its memory cost by 32$\times$ and also save energy thanks to the corresponding memory traffic reduction.
        
    \subsection{Weight Gradient Quantization}
    
        In common with other BNN training approaches, we employ ``straight-through estimation'' (STE) to facilitate gradient propagation in the presence of discretization in forward functions.
        STE approximates the gradient of a discontinuity by disregarding the derivative of the discretizer itself.
		As shown in Table~\ref{tab:memory_footprint}, {\tt float32} gradients were typically used with STE in the past.
        Intuitively, BNNs should be particularly robust to weight gradient quantization since their weights only constitute signs.
        On line~\ref{alg:bnn_ours_bin_dw} of Algorithm~\ref{alg:bnn_ours}, therefore, we binarize and store post-STE weight gradients, $\partial\hat{\boldsymbol{W}}$, for weight update.
        During that phase, we attenuate the gradients by $\sqrt{N_l}$, where $N_l$ is layer $l$'s fan-in, to reduce the learning rate and prevent premature weight clipping as advised by Sari \emph{et al.}~\cite{sari_how_does_bn_help_bnn} (line~\ref{alg:bnn_ours_opt_w}).
        Since fully connected layers are used as an example in Algorithm~\ref{alg:bnn_ours}, $N_l = M_{l-1}$ in this instance.
        
        Table~\ref{tab:memory_footprint} shows that, with binarization, the portion of our exemplary training run's memory consumption attributable to weight gradients dropped from 53.49 to just 1.67~MiB, leaving the scarce resources available for more quantization-sensitive variables such as $\boldsymbol{W}$ and momenta.
        Energy consumption will also decrease due to the associated reduction in memory traffic.
        
\section{Evaluation}

    \subsection{Keras Emulation}

        We built a GPU-based implementation emulating our BNN training method using Keras and TensorFlow, and experimented with the small-scale MNIST, CIFAR-10, and SVHN datasets, as well as large-scale ImageNet, using a range of network models.
        By emulating our algorithm on GPU, we can leverage the many powerful ML training softwares developed around it, and obtain large batches of experimental results in a short period of time.
        Our emulation environment is built on a Nvidia GeForce RTX 3090 GPU cluster with Red Hat Linux 9 operating system.
        Our baseline for comparison was the standard BNN training method introduced by Courbariaux \& Bengio~\cite{BNN_CNN_BinaryNet}, and we followed those authors' practice of reporting the highest test accuracy achieved in each run.
        Note that we did not tune hyperparameters, thus it is likely that higher accuracy than we report is achievable.
    
        \subsubsection{Small-Scale Datasets}
        
            For MNIST we evaluated using a five-layer MLP---henceforth simply denoted ``MLP''---with 256 neurons per hidden layer, and CNV~\cite{BNN_CNN_FINN} and BinaryNet~\cite{BNN_CNN_BinaryNet} for both CIFAR-10 and SVHN.
            We used three popular BNN optimizers: Adam~\cite{ADAM}, stochastic gradient descent (SGD) with momentum, and Bop~\cite{BOP}.
            While all three function reliably with our training scheme, we used Adam by default due to its stability.
            We used the development-based learning rate scheduling approach proposed by Wilson \emph{et al.}~\cite{marginal_value_of_adaptive_gradient} with an initial learning rate $\eta$ of 0.001 for all optimizers except for SGD with momentum, for which we used 0.1.
            We used batch size $B = 100$ for all except for Bop, for which we used $B = 50$ as recommended by Helwegen \emph{et al.}~\cite{BOP}.
            MNIST and CIFAR-10 were trained for 1000 epochs; SVHN for 200.
        
            Our choice of quantization targets primarily rested on the intuition that BNNs should be more robust to approximation in backward propagation than their higher-precision counterparts.
            To illustrate that this is indeed the case, we applied our method to both BNNs and {\tt float32} networks, with identical topologies and hyperparameters.
            Results of those experiments are shown in Table~\ref{app:tab:bnn_robust_to_quant}, in which significantly higher accuracy degradation was observed for the non-binary networks, as expected.
    
            \begin{table*}
            	\centering
            	\caption{
            	    Test accuracy of non-binary networks and BNNs using the standard and our proposed training approaches with Adam and a batch size of 100.
                    Results for our training approach applied to the former are included for reference only; we do not advocate for its use with non-binary networks.
                }
                \begin{threeparttable}
            	    \begin{tabular}{ccSSSSS>{\bf}S>{\bf}S}
            			\toprule
            			\multirow{3}[3]{*}{Model} & \multirow{3}[3]{*}{Dataset} & \multicolumn{7}{c}{Top-1 test accuracy} \\
            			\cmidrule(lr){3-9}
            			 &  & \multicolumn{3}{c}{Standard training} & \multicolumn{2}{c}{Reference training} & \multicolumn{2}{c}{\bf Proposed training} \\
            			 \cmidrule(lr){3-5} \cmidrule(lr){6-7} \cmidrule(lr){8-9}
            			 &  & {NN (\%)\tnote{1}} & {BNN (\%)} & {$\Delta$ (pp)} & {NN (\%)\tnote{1}} & {$\Delta$ (pp)}\tnote{2} & {BNN (\%)} & {$\Delta$ (pp)}\normalfont\tnote{3} \\
            			\midrule
            			MLP~\cite{BNN_CNN_FINN} & MNIST & 98.22 & 98.24 & 0.02 & 89.98 & -8.24 & 96.90 & -1.34 \\
            			CNV~\cite{BNN_CNN_FINN} & CIFAR-10 & 86.37 & 82.67 & -3.70 & 69.88 & -16.49 & 83.08 & 0.41 \\
            			CNV & SVHN & 97.30 & 96.37 & -0.93 & 79.44 & -17.86 & 94.28 & -2.09 \\
            			BinaryNet~\cite{BNN_CNN_BinaryNet} & CIFAR-10 & 88.20 & 88.74 & 1.61 & 76.56 & -11.64 & 89.09 & 0.35 \\
            			BinaryNet & SVHN & 96.54 & 97.40 & 0.86 & 85.71 & -10.83 & 95.93 & -1.47 \\
            			\bottomrule
            		\end{tabular}
            		\begin{tablenotes}
            		    \footnotesize
            		    \item[1] Non-binary neural network.
            		    \item[2] Baseline: non-binary network with standard training.
            		    \item[3] Baseline: BNN with standard training.
            		\end{tablenotes}
            	\end{threeparttable}
            	\label{app:tab:bnn_robust_to_quant}
            \end{table*}
    
            While our proposed BNN training method does exhibit limited accuracy degradation, as can be seen for three cases in Table~\ref{tab:benchmark_acc_mem_energy}, this comes in return for a geomean modeled memory saving of 3.67$\times$.
            It is also interesting to note that the reduction achievable for a given dataset depends on the model used.
            This observation is largely orthogonal to our work: by applying our approach to the training of a more appropriately chosen model, one can obtain the advantages of both optimized network selection and training.
        
        	\begin{table}
        	    \centering
        	    \caption{Test accuracy and memory footprint of the standard and our proposed training schemes using Adam and a batch size of 100.}
        	    \begin{tabular}{cS>{\bf}S>{\bf}SS>{\bf}S>{\bf}S}
        			\toprule
        	        \multirow{3}[1]{*}{\makecell{Model\\(Dataset)}} & \multicolumn{3}{c}{Top-1 test accuracy} & \multicolumn{3}{c}{Modeled memory} \\
        	        \cmidrule(lr){2-4} \cmidrule(lr){5-7}
        	         & {\multirow{2}{*}{\makecell{Std.\\(\%)}}} & {\multirow{2}{*}{\makecell{Prop.\\(\%)}}} & {\multirow{2}{*}{\makecell{$\Delta$\\(pp)}}} 
        	         & {\multirow{2}{*}{\makecell{Std.\\(MiB)}}} & {\multirow{2}{*}{\makecell{Prop.\\(MiB)}}} & {\multirow{2}{*}{\makecell{$\Delta$\\($\times\downarrow$)}}} \\
        	        \\
        	        \midrule
        	        \makecell{MLP\\(MNIST)} & 98.24 & 96.90 & -1.34 & 7.40 & 2.65 & 2.78 \\
        	        \makecell{CNV\\(CIFAR-10)} & 82.67 & 83.08 & 0.41 & 134.05 & 32.16 & 4.17 \\
        	        \makecell{CNV\\(SVHN)} & 96.37 & 94.28 & -2.09 & 134.05 & 32.16 & 4.17 \\
        	        \makecell{BinaryNet\\(CIFAR-10)} & 88.74 & 89.09 & 0.35 & 512.81 & 138.15 & 3.71 \\
        	        \makecell{BinaryNet\\(SVHN)} & 97.40 & 95.93 & -1.47 & 512.81 & 138.15 & 3.71 \\
        	        \bottomrule
        	    \end{tabular}
        	    \label{tab:benchmark_acc_mem_energy}
        	\end{table}
        	
            In order to explore the impacts of the various facets of our scheme, we applied them sequentially while training BinaryNet to classify CIFAR-10 with multiple optimizers.
         	As shown in Table~\ref{tab:binarynet_quant}, choices of data type, optimizer, and batch normalization implementation lead to tradeoffs against performance and memory costs.
            Major savings are attributable to the use of {\tt float16} variables and through the high-precision activation elimination our $\ell_1$ norm-based batch normalization facilitates.
            
            \begin{table*}
        		\centering
        		\caption{
        		    Impacts of moving from the standard to our proposed data representations with BinaryNet and CIFAR-10 and a batch size of 100.
                }
                \begin{threeparttable}
        		    \begin{tabular}{ccccSSSS}
        				\toprule
        				\multirow{2}[1]{*}{Optimizer} &
        				\multicolumn{2}{c}{Data type} & \multirow{2}[1]{*}{\makecell{Batch\\normalization}} & \multicolumn{2}{c}{Top-1 test accuracy} & \multicolumn{2}{c}{Modeled memory} \\
        				\cmidrule(lr){2-3} \cmidrule(lr){5-6} \cmidrule(lr){7-8}
        				 & $\partial\boldsymbol{W}$ & $\partial\boldsymbol{Y}$ &  & {\%} & {$\Delta$ (pp)}\tnote{1} & {MiB} & {$\Delta$ ($\times\downarrow$)}\tnote{1} \\
        				\midrule
        				 & {\tt float32} & {\tt float32} & $\ell_2$ & 88.74 & {--} & 512.81 & {--} \\
        				 & {\tt float16} & {\tt float16} & $\ell_2$ & 88.71 & -0.03 & 256.41 & 2.00 \\
        				Adam  & {\tt bool} & {\tt float16} & $\ell_2$ & 87.93 & -0.81 & 231.33 & 2.22 \\
        				 & {\tt bool} & {\tt float16} & $\ell_1$ & 89.69 & 0.95 & 231.33 & 2.22 \\
        				 & \bf{\tt bool} & \bf{\tt float16} & Proposed & 89.09 & 0.35 & 138.15 & 3.71 \\
        				\midrule
        				 & {\tt float32} & {\tt float32} & $\ell_2$ & 88.52 & {--} & 459.32 & {--} \\
        				\multirow{3}{*}{\makecell{SGD with\\momentum}} & {\tt float16} & {\tt float16} & $\ell_2$ & 88.54 & 0.02 & 229.66 & 2.00 \\
        				 & {\tt bool} & {\tt float16} & $\ell_2$ & 87.35 & -1.17 & 204.58 & 2.25 \\
        				 & {\tt bool} & {\tt float16} & $\ell_1$ & 89.09 & 0.57 & 204.58 & 2.25 \\
        				 & \bf{\tt bool} & \bf{\tt float16} & Proposed & 88.10 & -0.42 & 109.20 & 4.21 \\
        				\midrule
        				 & {\tt float32} & {\tt float32} & $\ell_2$ & 91.38 & {--} & 405.83 & {--} \\
        				 & {\tt float16} & {\tt float16} & $\ell_2$ & 91.36 & -0.02 & 202.92 & 2.00 \\
        				Bop  & {\tt bool} & {\tt float16} & $\ell_2$ & 90.54 & -0.84 & 177.84 & 2.28 \\
        				 & {\tt bool} & {\tt float16} & $\ell_1$ & 91.27 & -0.11 & 177.84 & 2.28 \\
        				 & \bf{\tt bool} & \bf{\tt float16} & Proposed & 91.48 & 0.10 & 82.45 & 4.92\\
        				\bottomrule
        			\end{tabular}
        			\begin{tablenotes}
        			    \footnotesize
        			    \item[1] Baseline: {\tt float32} $\partial\boldsymbol{W}$ and $\partial\boldsymbol{X}$ with standard ($\ell_2$) batch normalization.
        			\end{tablenotes}
        		\end{threeparttable}
        		\label{tab:binarynet_quant}
        	\end{table*}
            
            Fig.~\ref{plot:memory_saving} shows the modeled memory footprint savings from our proposed BNN training method for different optimizers and batch sizes, again for BinaryNet with the CIFAR-10 dataset.
            Across all of these, we achieved a geomean reduction of 4.81$\times$.
            Also observable from Fig.~\ref{plot:memory_saving} is that, for all optimizers, movement from the standard to our proposed BNN training allows the batch size used to increase by around 10$\times$, facilitating faster completion, without a material memory increase.
            Fig.~\ref{plot:memory_saving} finally shows that test accuracy does not drop significantly due to our approximations.
            With Adam and Bop, accuracy was near-identical, while with SGD we actually saw modest improvements.
            Unlike Adam or Bop, the standard SGD optimizer is unable to adapt its learning rate during gradient descent, thus scaling in batch size means scaling in learning rate.
            This leads to the decline in accuracy we see in Fig.~\ref{plot:memory_saving_sgd}, where increasing the batch size leads to undesirable learning rates.
            Our method, on the other hand, binarizes the weight gradients, which effectively normalizes the learning rate from the effects of batch size scaling.
            
            \begin{figure*}
                \centering
                \begin{tikzpicture}
    
    \begin{groupplot}[
        scale only axis,
		width=0.28\textwidth,
		height=0.28\textwidth,
		enlarge x limits=0.2,
		group style={group size=3 by 1, ylabels at=edge left, yticklabels at=edge left, xticklabels at=edge bottom, horizontal sep=1em},
        ybar,
        axis y line*=left,
        ymin=0,
        ymax=4,
        xtick={1, 2, 3, 4},
        xtick align=inside,
        xticklabels={100, 200, 500, 1000},
        xlabel near ticks,
        ylabel near ticks,
        ylabel= Modeled memory (GiB),
        /pgf/bar width=12pt,
        legend image code/.code={
            \draw[#1, bar width=6pt, yshift=-0.3em] plot coordinates {(0cm,0.8em)};
        }
    ]
            
        \nextgroupplot
        
        \node [text width=10em, anchor=north] at (axis description cs:0.5, 1) {\subcaption{Adam\label{plot:memory_saving_adam}}};
        
        \addplot [thick, pattern=north east lines, pattern color=black!25!cyan] table [x=id, y=baseline_adam_mem] {data/memory_footprint.txt};
        \label{plt:mem_footprint_baseline};
        \addplot [thick, pattern=north west lines, pattern color=orange] table [x=id, y=ours_adam_mem] {data/memory_footprint.txt}
        node [pos=0, xshift=0.6*\pgfkeysvalueof{/pgf/bar width}, rotate=90, anchor=west] {\small 3.71$\times$}
        node [pos=0.33, xshift=0.6*\pgfkeysvalueof{/pgf/bar width}, rotate=90, anchor=west] {\small 4.27$\times$}
        node [pos=0.67, xshift=0.6*\pgfkeysvalueof{/pgf/bar width}, rotate=90, anchor=west] {\small 4.85$\times$}
        node [pos=1, xshift=0.6*\pgfkeysvalueof{/pgf/bar width}, rotate=90, anchor=west] {\small 5.15$\times$};
        \label{plt:mem_footprint_ours};
        
        \nextgroupplot[
            xlabel=Batch size
        ]
        
        \node [text width=10em, anchor=north] at (axis description cs:0.5, 1) {\subcaption{SGD with momentum\label{plot:memory_saving_sgd}}};
        
        \addplot [thick, pattern=north east lines, pattern color=black!25!cyan] table [x=id, y=baseline_sgd_mem] {data/memory_footprint.txt};
        \addplot [thick, pattern=north west lines, pattern color=orange] table [x=id, y=ours_sgd_mem] {data/memory_footprint.txt}
        node [pos=0, xshift=0.6*\pgfkeysvalueof{/pgf/bar width}, rotate=90, anchor=west] {\small 4.21$\times$}
        node [pos=0.33, xshift=0.6*\pgfkeysvalueof{/pgf/bar width}, rotate=90, anchor=west] {\small 4.64$\times$}
        node [pos=0.67, xshift=0.6*\pgfkeysvalueof{/pgf/bar width}, rotate=90, anchor=west] {\small 5.09$\times$}
        node [pos=1, xshift=0.6*\pgfkeysvalueof{/pgf/bar width}, rotate=90, anchor=west] {\small 5.29$\times$};
        
        \nextgroupplot
        
        \node [text width=10em, anchor=north] at (axis description cs:0.5, 1) {\subcaption{Bop\label{plot:memory_saving_bop}}};
        
        \addplot [thick, pattern=north east lines, pattern color=black!25!cyan] table [x=id, y=baseline_bop_mem] {data/memory_footprint.txt};
        \addplot [thick, pattern=north west lines, pattern color=orange] table [x=id, y=ours_bop_mem] {data/memory_footprint.txt}
        node [pos=0, xshift=0.6*\pgfkeysvalueof{/pgf/bar width}, rotate=90, anchor=west] {\small 4.92$\times$}
        node [pos=0.33, xshift=0.6*\pgfkeysvalueof{/pgf/bar width}, rotate=90, anchor=west] {\small 5.16$\times$}
        node [pos=0.67, xshift=0.6*\pgfkeysvalueof{/pgf/bar width}, rotate=90, anchor=west] {\small 5.36$\times$}
        node [pos=1, xshift=0.6*\pgfkeysvalueof{/pgf/bar width}, rotate=90, anchor=west] {\small 5.44$\times$};   

    \end{groupplot}

    \begin{groupplot}[
        axis x line=none,
        axis y line*=right,
        scale only axis,
		width=0.28\textwidth,
		height=0.28\textwidth,
		enlarge x limits=0.2,
		group style={group size=3 by 1, ylabels at=edge right, yticklabels at=edge right, horizontal sep=1em},
        ylabel near ticks,
        ylabel=Top-1 test accuracy (\%),
        ymin=81,
        ymax=95,
        error bars/y dir=both,
        error bars/y explicit=true,
    ]
        
        \nextgroupplot 
        
        \addplot [thick, only marks, mark=o, mark options={scale=1.5}] table [x=id, y=baseline_adam_acc, y error plus=baseline_adam_errU, y error minus=baseline_adam_errL] {data/memory_footprint.txt};
        \label{plt:mem_footprint_baseline_acc}
        \addplot [thick, only marks, mark=triangle, mark options={scale=1.5}] table [x=id, y=ours_adam_acc, y error plus=ours_adam_errU, y error minus=ours_adam_errL] {data/memory_footprint.txt};
        \label{plt:mem_footprint_ours_acc}
        
        \nextgroupplot 
        
        \addplot [thick, only marks, mark=o, mark options={scale=1.5}] table [x=id, y=baseline_sgd_acc, y error plus=baseline_sgd_errU, y error minus=baseline_sgd_errL] {data/memory_footprint.txt};
        \addplot [thick, only marks, mark=triangle, mark options={scale=1.5}] table [x=id, y=ours_sgd_acc, y error plus=ours_sgd_errU, y error minus=ours_sgd_errL] {data/memory_footprint.txt};
        
        \nextgroupplot
        
        \addplot [thick, only marks, mark=o, mark options={scale=1.5}] table [x=id, y=baseline_bop_acc, y error plus=baseline_bop_errU, y error minus=baseline_bop_errL] {data/memory_footprint.txt};
        \addplot [thick, only marks, mark=triangle, mark options={scale=1.5}] table [x=id, y=ours_bop_acc, y error plus=ours_bop_errU, y error minus=ours_bop_errL] {data/memory_footprint.txt};
        
    \end{groupplot}
    
\end{tikzpicture}
                \adjustbox{margin=0.5em}{\begin{tikzpicture}
    
    \matrix[
        matrix of nodes,
        draw,
        inner sep=0.2em
    ] {
        \ref{plt:mem_footprint_baseline} & Memory (standard) &[0.5em]
        \ref{plt:mem_footprint_ours} & Memory (proposed) &[0.5em]
        \ref{plt:mem_footprint_baseline_acc} & Accuracy (standard) &[0.5em]
        \ref{plt:mem_footprint_ours_acc} & Accuracy (proposed) \\
    };

\end{tikzpicture}}
            	\caption{
            	    Batch size \emph{vs} training memory footprint and achieved test accuracy for BinaryNet with CIFAR-10.
                    Annotations show memory reductions for the proposed training approach.
                    Each test accuracy point marks the mean of five independent training runs, with an error bar indicating its distribution.
                }
            	\label{plot:memory_saving}
            \end{figure*}
            
            While not of concern with regards to memory consumption, decreases in convergence rate are undesirable due to their elongation of training times and, consequently, reduction of energy efficiency.
            In order to ensure that our algorithmic modifications do not cause material convergence rate degradation, we inspected the validation accuracy curves obtained during our training runs.
            Figs.~\ref{plot:training_curves_all_benchmakrs} and \ref{plot:training_curves} exemplify these for the experiments whose results were reported in Table~\ref{tab:benchmark_acc_mem_energy} and Fig.~\ref{plot:memory_saving}, respectively.
            No discernible change in convergence rate can be seen in any of the plots, thus we can be confident that our proposals will not negatively impact training times.
            
            \begin{figure*}
                \centering
                \begin{tikzpicture}
    
    \begin{groupplot} [
        scale only axis,
		group style={group size=3 by 2, ylabels at=edge left, yticklabels at=edge left, horizontal sep=1em, vertical sep=3em},
		width=0.27\textwidth,
		height=0.27\textwidth,
		xmin=0,
 		xmax=1000,
 		ymin=0.35,
 		ymax=1.2,
		xlabel near ticks,
        ylabel near ticks,
        yticklabel={\pgfmathparse{\tick*100}\pgfmathprintnumber{\pgfmathresult}},
        x tick label style={rotate=60, anchor=east, /pgf/number format/set thousands separator={}},
        max space between ticks=30em
	]
	
	    \nextgroupplot [
	        ylabel={Top-1 validation accuracy (\%)},
	        every axis y label/.append style={at=(ticklabel cs:0), xshift=-1.5em},
	    ]
	    
	    \node [text width=10em, anchor=north] at (axis description cs:0.5, 1) {\subcaption*{MLP/MNIST}};
        
        \addplot [thick, red] table [y=bnn_baseline, x=epoch] {data/training_curves_all_benchmarks/mlp_mnist.txt};
        \addplot [thick, black!25!green, densely dashed] table [y=bnn_ours, x=epoch] {data/training_curves_all_benchmarks/mlp_mnist.txt};
        
	    \nextgroupplot
	    
	    \node [text width=10em, anchor=north] at (axis description cs:0.5, 1) {\subcaption*{CNV/CIFAR-10}};
	    
	    \addplot [thick, red] table [y=bnn_baseline, x=epoch] {data/training_curves_all_benchmarks/cnv_cifar.txt};
        \addplot [thick, black!25!green, densely dashed] table [y=bnn_ours, x=epoch] {data/training_curves_all_benchmarks/cnv_cifar.txt};
        
	    \nextgroupplot [
	        xmax=200,
	        legend pos=south east,
	        legend entries={
	            Standard,
	            Proposed
	        }
	    ]
	    
	    \node [text width=10em, anchor=north] at (axis description cs:0.5, 1) {\subcaption*{CNV/SVHN}};
	    
	    \addplot [thick, red] table [y=bnn_baseline, x=epoch] {data/training_curves_all_benchmarks/cnv_svhn.txt};
        \addplot [thick, black!25!green, densely dashed] table [y=bnn_ours, x=epoch] {data/training_curves_all_benchmarks/cnv_svhn.txt};
        
        \nextgroupplot [
	        xshift=0.135\textwidth+0.5em,
	        xlabel={Epoch},
            every axis x label/.append style={at=(ticklabel cs:1), xshift=0.5em}        ]
        
        \node [text width=10em, anchor=north] at (axis description cs:0.5, 1) {\subcaption*{BinaryNet/CIFAR-10}};
        
        \addplot [thick, red] table [y=bnn_baseline, x=epoch] {data/training_curves_all_benchmarks/binarynet_cifar.txt};
        \addplot [thick, black!25!green, densely dashed] table [y=bnn_ours, x=epoch] {data/training_curves_all_benchmarks/binarynet_cifar.txt};
        
        \nextgroupplot [
	        xshift=0.135\textwidth+0.5em,
	        xmax=200
        ]
        
        \node [text width=10em, anchor=north] at (axis description cs:0.5, 1) {\subcaption*{BinaryNet/SVHN}};
        
        \addplot [thick, red] table [y=bnn_baseline, x=epoch] {data/training_curves_all_benchmarks/binarynet_svhn.txt};
        \addplot [thick, black!25!green, densely dashed] table [y=bnn_ours, x=epoch] {data/training_curves_all_benchmarks/binarynet_svhn.txt};
        
        
    
    \end{groupplot}

\end{tikzpicture}
                \caption{
                Comparison in achieved validation accuracy curves between the standard and our proposed training schemes with multiple combinations in models and datasets, using Adam and a batch size of 100.
                These plots correspond to results which are reported in Table~\ref{tab:benchmark_acc_mem_energy}.}
                \label{plot:training_curves_all_benchmakrs}
            \end{figure*}
            
            
            \begin{figure*}
                \centering
                \begin{tikzpicture}
    
    \begin{groupplot} [
        scale only axis,
		group style={group size=3 by 3, ylabels at=edge left, xticklabels at=edge bottom, yticklabels at=edge left, horizontal sep=1em, vertical sep=1em},
		width=0.27\textwidth,
		height=0.27\textwidth,
		xmin=0,
		xmax=1000,
		ymin=0.35,
		ymax=1.15,
		xlabel near ticks,
        ylabel near ticks,
        x tick label style={rotate=60, anchor=east, /pgf/number format/set thousands separator={}},
        yticklabel={\pgfmathparse{\tick*100}\pgfmathprintnumber{\pgfmathresult}},
        max space between ticks=30em
	]
	    \nextgroupplot
	    
	    \node [text width=20em, anchor=north] at (axis description cs:0.5, 1) {\subcaption*{Adam\\($B=100$)}};
        
        \addplot [thick, red] table [y=bnn_baseline, x=epoch] {data/training_curves/adam_100.txt}; \label{plt:training_curve_adam_100_baseline}
        \addplot [thick, black!25!green, densely dashed] table [y=bnn_ours, x=epoch] {data/training_curves/adam_100.txt}; \label{plt:training_curve_adam_100_ours}
        
	    \nextgroupplot
	    
	    \node [text width=20em, anchor=north] at (axis description cs:0.5, 1) {\subcaption*{SGD with momentum\\($B=100$)}};
	    
	    \addplot [thick, red] table [y=bnn_baseline, x=epoch] {data/training_curves/sgd_100.txt};
        \addplot [thick, black!25!green, densely dashed] table [y=bnn_ours, x=epoch] {data/training_curves/sgd_100.txt};
        
	    \nextgroupplot [
	        legend pos=south east,
	        legend entries={Standard, Proposed}
	    ]
	    
	    \node [text width=20em, anchor=north] at (axis description cs:0.5, 1) {\subcaption*{Bop\\($B=100$)}};
	    
	    \addplot [thick, red] table [y=bnn_baseline, x=epoch] {data/training_curves/bop_100.txt};
        \addplot [thick, black!25!green, densely dashed] table [y=bnn_ours, x=epoch] {data/training_curves/bop_100.txt};
        
        \nextgroupplot [
            ylabel={Top-1 validation accuracy (\%)}
        ]
        
        \node [text width=20em, anchor=north] at (axis description cs:0.5, 1) {\subcaption*{Adam\\($B=500$)}};
        
        \addplot [thick, red] table [y=bnn_baseline, x=epoch] {data/training_curves/adam_500.txt};
        \addplot [thick, black!25!green, densely dashed] table [y=bnn_ours, x=epoch] {data/training_curves/adam_500.txt};
        
	    \nextgroupplot
	    
	    \node [text width=20em, anchor=north] at (axis description cs:0.5, 1) {\subcaption*{SGD with momentum\\($B=500$)}};
        
        \addplot [thick, red] table [y=bnn_baseline, x=epoch] {data/training_curves/sgd_500.txt};
        \addplot [thick, black!25!green, densely dashed] table [y=bnn_ours, x=epoch] {data/training_curves/sgd_500.txt};
        
	    \nextgroupplot
	    
	    \node [text width=20em, anchor=north] at (axis description cs:0.5, 1) {\subcaption*{Bop\\($B=500$)}};
        
        \addplot [thick, red] table [y=bnn_baseline, x=epoch] {data/training_curves/bop_500.txt};
        \addplot [thick, black!25!green, densely dashed] table [y=bnn_ours, x=epoch] {data/training_curves/bop_500.txt};
        
        \nextgroupplot
        
        \node [text width=20em, anchor=north] at (axis description cs:0.5, 1) {\subcaption*{Adam\\($B=1000$)}};
        
        \addplot [thick, red] table [y=bnn_baseline, x=epoch] {data/training_curves/adam_1000.txt};
        \addplot [thick, black!25!green, densely dashed] table [y=bnn_ours, x=epoch] {data/training_curves/adam_1000.txt};
        
	    \nextgroupplot [
	        xlabel={Epoch}
	    ]
	    
	    \node [text width=20em, anchor=north] at (axis description cs:0.5, 1) {\subcaption*{SGD with momentum\\($B=1000$)}};
        
        \addplot [thick, red] table [y=bnn_baseline, x=epoch] {data/training_curves/sgd_1000.txt};
        \addplot [thick, black!25!green, densely dashed] table [y=bnn_ours, x=epoch] {data/training_curves/sgd_1000.txt};
        
	    \nextgroupplot
	    
	    \node [text width=20em, anchor=north] at (axis description cs:0.5, 1) {\subcaption*{Bop\\($B=1000$)}};
        
        \addplot [thick, red] table [y=bnn_baseline, x=epoch] {data/training_curves/bop_1000.txt};
        \addplot [thick, black!25!green, densely dashed] table [y=bnn_ours, x=epoch] {data/training_curves/bop_1000.txt};
    
    \end{groupplot}

\end{tikzpicture}
                \caption{Comparison in achieved validation accuracy curves between the standard and our proposed training schemes with multiple combinations in optimizers and batch sizes $\left(B\right)$, using BinaryNet model and CIFAR-10 dataset.
                These plots correspond to results which are reported in Fig.~~\ref{plot:memory_saving}.}
                \label{plot:training_curves}
            \end{figure*}
            
        \ew{Added paragraphs discussing NAS results here.}
		\jd{I removed the ``can reach the performance of float32 models'' sentence because that statement is redundant, given some of the intro material in the article. Perhaps you meant to make a different point (?).}
		\ew{Not really. I agree with the removal of the sentence.}
		\jd{I also wasn't really sure where you were going re overparameterisation of BNNs vs non-BNNs (since that's not what's being compared here), so I just took that sentence out. Do you still think we need it/something similar?}
		\ew{I agree with your proposed change.}
		For the results presented thus far, we made use of off-the-shelf network models.
		As confirmed by Zhang~\emph{et al.}, a network possess perfect expressivity once its number of parameters matches the number of data points used for its training~\cite{rethinking_generalization}.
		Consequently, most practical networks are overparameterized.
		While the impact of overparametrization on network generalization is an active research field~\cite{chatterjee2022generalization} and outside the scope of this work, we sought to investigate whether overparametrization was the source of robustness to gradient approximation that we observed of BNNs.
		To do this, we performed neural architecture search (NAS) for the MNIST, CIFAR-10 and SVHN datasets, comparing the impact of removing network redundancy on both the standard and our training approaches.
		We adopted Shen~\emph{et al.}'s approach to BNN NAS, applying it to the MLP and BinaryNet models as starting points~\cite{BNN_NAS}.
		Following their proposals, we set accuracy-to-parameter weight factor $\lambda$ to 0.1 for MLP with MNIST and 0.01 for BinaryNet with CIFAR-10 and SVHN.
		As shown in Table~\ref{tab:bnn_nas}, we achieved sizeable parameter reductions for all of these and, most importantly, observed no difference in accuracy degradation for the two training approaches.
		These experiments therefore suggest that the reduction of network complexity impacts both methods equally, and that the performance of ours is not reliant on overparameterization.
        
		\begin{table*}
    	    \centering
    	    \caption{Model complexity and test accuracy impacts of NAS under the standard and proposed training schemes.}
    	    \begin{tabular}{ccSSSSSS>{\bf}S>{\bf}S>{\bf}S}
    			\toprule
    	        \multirow{3}[2]{*}{Model} & \multirow{3}[2]{*}{Dataset} & \multicolumn{3}{c}{Parameters (M)} & \multicolumn{6}{c}{Top-1 test accuracy} \\
				\cmidrule(lr){3-5} \cmidrule(lr){6-11}
				 & & {Pre} & \multicolumn{2}{c}{Post} & \multicolumn{3}{c}{Standard training} & \multicolumn{3}{c}{\bf Proposed training} \\
				\cmidrule(lr){3-3} \cmidrule(lr){4-5} \cmidrule(lr){6-8} \cmidrule(lr){9-11}
				 & & {\#} & {\#} & {\makecell{$\Delta$\\($\times\downarrow$)}} & {\makecell{Pre\\(\%)}} & {\makecell{Post\\(\%)}} & {\makecell{$\Delta$\\(pp)}} & {\makecell{Pre\\(\%)}} & {\makecell{Post\\(\%)}} & {\makecell{$\Delta$\\(pp)}} \\
				\midrule
				MLP & MNIST & 0.40 & 0.16 & 2.52 & 98.24 & 97.58 & -0.66 & 96.90 & 96.35 & -0.55 \\
				BinaryNet & CIFAR-10 & 14.02 & 3.62 & 3.87 & 88.74 & 87.14 & -1.60 & 89.09 & 87.17 & -1.92 \\
				BinaryNet & SVHN & 14.02 & 3.77 & 3.72 & 97.40 & 97.26 & -0.14 & 95.93 & 95.38 & -0.55 \\
				\bottomrule
    	    \end{tabular}
    	    \label{tab:bnn_nas}
    	\end{table*}
        
        \subsubsection{ImageNet}
        
            We also trained ResNetE-18~\cite{back_to_simplicity_resnete_18} and Bi-Real-18~\cite{birealnet}---mixed-precision models with most convolutional layers binarized---to classify ImageNet.
            These models are representative of a broad class of ImageNet-capable networks, thus similar results should be achievable for others with which they share architectural features.
            Finding development-based learning rate scheduling to not work well with ResNetE-18, we resorted to the fixed decay schedule described by Bethge \emph{et al.}~\cite{back_to_simplicity_resnete_18}.
            $\eta$ began at 0.016 and decayed by a factor of 10 at epochs 70, 90, and 110.
            We trained for 120 epochs with $B = 4096$.
            For Bi-Real-18, we trained for 80 epochs with $B = 512$ and a cosine-decaying learning rate starting from $\eta = 0.001$.
            Both models were optimized using Adam.
            
            We show the performance of these benchmarks when applying each of our proposed approximations in turn, as well their assemblage, in Table~\ref{tab:imagenet}.
            Since the Tensor Processing Units we used here natively support {\tt bfloat16} rather than {\tt float16}, we switched to the former for these experiments.
            Where {\tt bfloat16} variables were used, these were employed across all layers; the remaining approximations were applied only to binary layers.
            While these savings are smaller than those for our small-scale experiments, we note that the first convolutional layer of both ResNetE-18 and Bi-Real-18 is the largest and is non-binary, thus its activation storage dwarfs that of the remaining layers.
            We also remark that, while $\sim$2~pp accuracy drops may not be acceptable for some application deployments, sizable memory reductions are otherwise achievable.
            The effects of binarized $\partial\boldsymbol{W}$ are insignificant since ImageNet's large images result in proportionally small weight memory occupancy.
            
            \begin{table*}
        	    \centering
        	    \caption{Test accuracy and memory footprint of the standard and proposed schemes for ImageNet training with Adam and a batch size of 4096.}
        	    \begin{threeparttable}
            	    \begin{tabular}{cSSSSSSSS}
            			\toprule
            			 & \multicolumn{4}{c}{ResNetE-18} & \multicolumn{4}{c}{Bi-Real-18} \\
            			\cmidrule(lr){2-5} \cmidrule(lr){6-9}
            	        Approximations & \multicolumn{2}{c}{\makecell{Top-1\\test acc.}} & \multicolumn{2}{c}{\makecell{Modeled\\memory}} & \multicolumn{2}{c}{\makecell{Top-1\\test acc.}} & \multicolumn{2}{c}{\makecell{Modeled\\memory}} \\
            	        \cmidrule(lr){2-3} \cmidrule(lr){4-5} \cmidrule(lr){6-7} \cmidrule(lr){8-9}
            	         & {\%} & {$\Delta$ (pp)}\tnote{1} & {GiB} & {$\Delta$ ($\times\downarrow$)}\tnote{1} & {\%} & {$\Delta$ (pp)}\tnote{1} & {GiB} & {$\Delta$ ($\times\downarrow$)}\tnote{1} \\
            	        \midrule
            	        None & 58.77 & {--} & 70.11 & {--} & 56.71 & {--} & 70.11 & {--} \\
            	        All-{\tt bfloat16} & 58.85 & 0.08 & 35.45 & 1.98 & 56.72 & 0.01 & 35.45 & 1.98  \\
            	        {\tt bool} $\partial\boldsymbol{W}$ only & 57.59 & -1.28 & 70.07 & 1.00 & 55.69 & -1.02 & 70.07 & 1.00 \\
            	        $\ell_1$ batch norm. only & 58.34 & -0.43 & 70.11 & 1.00 & 56.08 & -0.63 & 70.11 & 1.00 \\
            	        Prop. batch norm. only & 58.23 & -0.54 & 47.86 & 1.46 & 55.59 & -1.12 & 47.86 & 1.46 \\
            	        \bf Proposed\normalfont & \bf 57.04 & \bf -1.73 & \bf 18.54 & \bf 3.78 & \bf 54.45 & \bf -2.26 & \bf 18.54 & \bf 3.78 \\
            	        \bottomrule
            	    \end{tabular}
            	    \begin{tablenotes}
                	    \footnotesize
            			\item[1] Baseline: approximation-free training.
                	\end{tablenotes}
            	\end{threeparttable}
        	    \label{tab:imagenet}
        	\end{table*}
            
            We acknowledge that dataset storage requirements likely render ImageNet training on edge platforms infeasible, and that network fine-tuning is a task more commonly deployed on devices of such scale.
            However, given that the accuracy changes and resource savings we report for more challenging, from-scratch training are favorable and reasonably consistent across a wide range of use-cases, we have confidence that positive results are readily achievable for fine-tuning as well.
            Nevertheless, our ImageNet proof of concept confirms the efficacy of large-scale neural network training on the edge.
            
            In common with our small-scale experiments, our proposals did not lead to noticeable convergence rate changes \emph{vs} the standard BNN training algorithm.
            This is evident from Fig.~\ref{plot:training_curves_imagenet}, which contains the validation accuracy curves obtained for the experiments whose results were reported in Table~\ref{tab:imagenet}.
            
            \begin{figure*}
                \centering
                \begin{tikzpicture}
    
    \begin{groupplot} [
        scale only axis,
		width=0.4\textwidth,
		height=0.4\textwidth,
		group style={group size=2 by 1, ylabels at=edge left, yticklabels at=edge left, xticklabels at=edge bottom, horizontal sep=1em},
 		xlabel near ticks,
        ylabel near ticks,
        ylabel={Top-1 validation accuracy (\%)},
        ymin=0.05,
        ymax=0.7,
        tick label style={/pgf/number format/fixed},
        x tick label style={rotate=60, anchor=east},
        yticklabel={\pgfmathparse{\tick*100}\pgfmathprintnumber{\pgfmathresult}},
        max space between ticks=30em
	]
            
        \nextgroupplot[
            xlabel={Epoch},
            every axis x label/.append style={at=(ticklabel cs:1), xshift=0.5em},
            xmin=0,
            xmax=120,
        ]
        
        \node [text width=20em, anchor=north] at (axis description cs:0.5, 1) {\subcaption*{ResNetE-18}};
        
        \addplot [thick, red] table [y=bnn_baseline, x=epoch_baseline] {data/training_curves_all_benchmarks/resnete18_imagenet.txt};
        \addplot [thick, black!25!green, densely dashed] table [y=bnn_float16, x=epoch_float16] {data/training_curves_all_benchmarks/resnete18_imagenet.txt};
        \addplot [thick, white!25!blue, dotted] table [y=bnn_bool_dw, x=epoch_bool_dw] {data/training_curves_all_benchmarks/resnete18_imagenet.txt};
        \addplot [thick, black!25!cyan, dashed] table [y=bnn_l1_bn, x=epoch_l1_bn] {data/training_curves_all_benchmarks/resnete18_imagenet.txt};
        \addplot [thick, black!25!magenta, loosely dotted] table [y=bnn_proposed_bn, x=epoch_proposed_bn] {data/training_curves_all_benchmarks/resnete18_imagenet.txt};
        \addplot [thick, violet, dashdotdotted] table [y=bnn_ours, x=epoch_ours] {data/training_curves_all_benchmarks/resnete18_imagenet.txt};
            
        \nextgroupplot[
            legend pos=south east,
            legend entries={
                None,
                All-{\tt bfloat16},
                {\tt bool} $\partial\boldsymbol{W}$ only,
                $l_1$ batch norm. only,
                Prop. batch norm. only,
                Final combination
            },
            xmin=0,
            xmax=80,
        ]
        
        \node [text width=20em, anchor=north] at (axis description cs:0.5, 1) {\subcaption*{Bi-Real-18}};
        
        \addplot [thick, red] table [y=bnn_baseline, x=epoch_baseline] {data/training_curves_all_benchmarks/bireal18_imagenet.txt};
        \addplot [thick, black!25!green, densely dashed] table [y=bnn_float16, x=epoch_float16] {data/training_curves_all_benchmarks/bireal18_imagenet.txt};
        \addplot [thick, white!25!blue, dotted] table [y=bnn_bool_dw, x=epoch_bool_dw] {data/training_curves_all_benchmarks/bireal18_imagenet.txt};
        \addplot [thick, black!25!cyan, dashed] table [y=bnn_l1_bn, x=epoch_l1_bn] {data/training_curves_all_benchmarks/bireal18_imagenet.txt};
        \addplot [thick, black!25!magenta, loosely dotted] table [y=bnn_proposed_bn, x=epoch_proposed_bn] {data/training_curves_all_benchmarks/bireal18_imagenet.txt};
        \addplot [thick, violet, dashdotdotted] table [y=bnn_ours, x=epoch_ours] {data/training_curves_all_benchmarks/bireal18_imagenet.txt};
    
    \end{groupplot}

\end{tikzpicture}
                \caption{Achieved validation accuracy over time for the experiments whose results are reported in Table~~\ref{tab:imagenet}.}
                \label{plot:training_curves_imagenet}
            \end{figure*}
    
    \subsection{Embedded Platform Prototypes}
    
        To more concretely demonstrate the benefits of our proposed training method, we also wrote software targeting an embedded-scale computing platform.
        We chose to use a Raspberry Pi 3B+, a popular single-board computer with hardware representative of current mobile and other edge devices, for this purpose.
        The platform features a four-core, 64-bit Arm Cortex-A53 CPU clocked at 1.4~GHz and 1~GiB of LPDDR2 RAM.
        We used the PyPI {\tt memory\_profiler} module and Valgrind to monitor the memory occupancy of Keras- and C++-based implementations, respectively.
        Energy consumption was logged with a standard USB power meter connected to the Raspbberry Pi's external power supply~\cite{RPI_POWER_METER}.
        
        \subsubsection{Na\"ive C++ Implementation}
        
            While existing training frameworks, including TensorFlow and PyTorch, allow for some data format customization, they lack support for direct control of variable storage.
            Moreover, when in training mode, they tend to reserve hundreds of MiBs of memory regardless of the model size, making their use infeasible on edge devices.
            TensorFlow-lite delivers low-memory inference, but it does not support training.
            Therefore, while these existing frameworks are useful for accuracy evaluation, implementations of our approach that realize its promised memory advantage must be built from scratch.
            Our first prototypes were direct implementations of Algorithms~\ref{alg:bnn_vanilla} and \ref{alg:bnn_ours} in C++.
            We also trained using Keras, where possible within the Raspberry Pi's memory limit, for comparison.
            
            Measurements of the peak memory use of our na\"ive C++ prototypes prove the validity of our memory model.
            As reflected in Fig.~\ref{plot:naive_prototype_memory}, two effects cause the model to produce underestimates.
            There is a constant, $\sim$5\% memory increase across all experiment pairs.
            This is attributable to process overheads, which we left unmodeled.
            There is also a second, batch size-correlated overhead due to activation copying between layers.
            This is significantly more pronounced for the standard algorithm due to its use of {\tt float32}---rather than {\tt bool}---activations.
            While we did not model these copies since they are not strictly necessary, their avoidance would have unbeneficially complicated our software.
            
            \begin{figure}
                \centering
                \begin{tikzpicture}
    
    \begin{axis}[
        scale only axis,
		width=0.62\textwidth,
		height=0.42\textwidth,
		enlarge x limits=0.2,
        ymin=0,
        ymax=30,
        xtick={1, 2, 3, 4},
        xticklabels={100, 200, 500, 1000},
        ylabel near ticks,
        /pgf/bar width=8pt,
        ybar,
	    ylabel=Memory (MiB),
	    xlabel=Batch size,
        xtick align=inside,
        legend image code/.code={
            \draw[#1, bar width=6pt, yshift=-0.3em] plot coordinates {(0cm,0.8em)};
        },
        legend pos=north west,
	    legend entries={Modeled (standard), Measured (standard), Modeled (proposed), Measured (proposed)}
    ]
        
        \addplot [thick, pattern=north east lines, pattern color=black!25!cyan] table [x=id, y=baseline_estimate] {data/naive_prototype_memory.txt};
        \label{plt:proto_mem_footprint_baseline};
        \addplot [thick, pattern=north west lines, pattern color=orange] table [x=id, y=baseline_measured] {data/naive_prototype_memory.txt}
        node [pos=0, xshift=-0.6*\pgfkeysvalueof{/pgf/bar width}, rotate=90, anchor=west] {\small 1.14$\times$}
        node [pos=0.33, xshift=-0.6*\pgfkeysvalueof{/pgf/bar width}, rotate=90, anchor=west] {\small 1.22$\times$}
        node [pos=0.67, xshift=-0.6*\pgfkeysvalueof{/pgf/bar width}, rotate=90, anchor=west] {\small 1.41$\times$}
        node [pos=1, xshift=-0.6*\pgfkeysvalueof{/pgf/bar width}, rotate=90, anchor=west] {\small 1.57$\times$};
        \label{plt:proto_mem_footprint_baseline_measured};
        \addplot [thick, pattern=vertical lines, pattern color=blue] table [x=id, y=ours_estimate] {data/naive_prototype_memory.txt};
        \label{plt:proto_mem_footprint_ours};
        \addplot [thick, pattern=horizontal lines, pattern color=red] table [x=id, y=ours_measured] {data/naive_prototype_memory.txt}
        node [pos=0, xshift=2.0*\pgfkeysvalueof{/pgf/bar width}, rotate=90, anchor=west] {\small 1.07$\times$}
        node [pos=0.33, xshift=2.0*\pgfkeysvalueof{/pgf/bar width}, rotate=90, anchor=west] {\small 1.07$\times$}
        node [pos=0.67, xshift=2.0*\pgfkeysvalueof{/pgf/bar width}, rotate=90, anchor=west] {\small 1.08$\times$}
        node [pos=1, xshift=2.0*\pgfkeysvalueof{/pgf/bar width}, rotate=90, anchor=west] {\small 1.09$\times$};
        \label{plt:proto_mem_footprint_ours_measured};

    \end{axis}
    
\end{tikzpicture}
            	\caption{
            	    Batch size \emph{vs} memory footprint for our na\"ive C++ prototypes training MLP to classify MNIST with Adam.
            	    Annotations mark the ratio between measured and modeled memory pairs.
                }
            	\label{plot:naive_prototype_memory}
            \end{figure}

			\jd{I'm having a harder time with this presentation than I did with the old one. I thought the point of the former Fig. 7(c) was to show the memory vs energy tradeoff. Presenting memory and energy separately (as now in Fig. 8) seems to prevent that. It's also unclear what's going on with the annotations below (inc. because they're not described in the caption).}
			\ew{I think the memory and energy do not necessary form a tradeoff, since savings in memory lead to savings in energy. Besides, we only have two data points, which are not enough to form any trend frontier. I have modified a sentence in the discussion to hopefully make the figure clearer to read.}
            \begin{figure*}
                \centering
                \begin{tikzpicture}
    
    \begin{groupplot}[
        scale only axis,
		width=0.35\textwidth,
		height=0.35\textwidth,
		enlarge x limits=0.1,
		enlarge y limits=0.1,
		group style={group size=2 by 1, ylabels at=edge left, horizontal sep=1em, yticklabels at=edge left},
        xmajorgrids=true,
        ymajorgrids=true,
        ylabel near ticks,
        xlabel near ticks,
        ymode=log,
        log ticks with fixed point,
        ymin=1,
        ymax=1000,
        xtick align=inside,
        /pgf/number format/1000 sep={}
    ]
            
        \nextgroupplot[
            xmode=log,
		    xlabel= Training time/batch (s),
            every axis x label/.append style={at=(ticklabel cs:1), xshift=0.5em},
		    ylabel=Measured memory (MiB)
        ]
        
        \node [text width=10em, anchor=north] at (axis description cs:0.5, 1) {\subcaption{MLP/MNIST\label{plot:prototype_mem_trainTime_mlp}}};
        
        \addplot [mark=triangle, mark options={scale=1.5}, thick, color=black!25!green] table [x=trainTime_keras, y=mem_keras] {data/prototype_measured/mlp_mem_trainTime_plot.txt};
        \label{plt:prototype_mlp_keras};
        \addplot [mark=star, mark options={scale=1.5}, thick, color=blue] table [x=trainTime_naiveStd, y=mem_naiveStd] {data/prototype_measured/mlp_mem_trainTime_plot.txt};
        \label{plt:prototype_mlp_naiveStd};
        \addplot [mark=+, mark options={scale=1.5}, thick, color=red] table [x=trainTime_naiveOurs, y=mem_naiveOurs] {data/prototype_measured/mlp_mem_trainTime_plot.txt};
        \label{plt:prototype_mlp_naiveOurs};
        \addplot [mark=o, mark options={scale=1.5}, thick, color=black!25!cyan] table [x=trainTime_optStd, y=mem_optStd] {data/prototype_measured/mlp_mem_trainTime_plot.txt};
        \label{plt:prototype_mlp_optStd};
        \addplot [mark=diamond, mark options={scale=1.5}, thick, color=orange] table [x=trainTime_optOurs, y=mem_optOurs] {data/prototype_measured/mlp_mem_trainTime_plot.txt};
        \label{plt:prototype_mlp_optOurs};
        
        \addplot [mark=triangle,
            mark size=4pt,
            mark options={
                draw=black,
                fill=white,
            },
            only marks,
            every mark/.append style={rotate=90},
        ]
        table {%
        0.2 1024
        }; \label{plt:rpi_max_mem}
        
        \nextgroupplot[
            xmode=log
        ]
        
        \node [text width=10em, anchor=north] at (axis description cs:0.5, 1) {\subcaption{BinaryNet/CIFAR-10\label{plot:prototype_mem_trainTime_binarynet}}};
        
        \addplot [mark=o, mark options={scale=1.5}, thick, color=black!25!cyan] table [x=trainTime_optStd, y=mem_optStd] {data/prototype_measured/binarynet_mem_trainTime_plot.txt};
        \label{plt:prototype_binarynet_optStd};
        \addplot [mark=diamond, mark options={scale=1.5}, thick, color=orange] table [x=trainTime_optOurs, y=mem_optOurs] {data/prototype_measured/binarynet_mem_trainTime_plot.txt};
        \label{plt:prototype_binarynet_optOurs};
        
        \node[anchor=south,draw,fill=white,inner sep=0.2em] at (axis description cs:0.5, 0.05) {\small
        \begin{tabular}{cl}
            \ref{plt:prototype_mlp_keras} & Standard (Keras)\\
            \ref{plt:prototype_mlp_naiveStd} & Standard (na\"ive C++)\\
            \ref{plt:prototype_mlp_optStd} & Standard (CBLAS)\\
            \ref{plt:prototype_mlp_naiveOurs} & Proposed (na\"ive C++)\\
            \ref{plt:prototype_mlp_optOurs} & Proposed (CBLAS)
        \end{tabular}};

    \end{groupplot}
    
\end{tikzpicture}
            	\caption{
            	    Measured peak memory consumption \emph{vs} training time \subref{plot:prototype_mem_trainTime_mlp}--\subref{plot:prototype_mem_trainTime_binarynet} per batch for implementations training MLP/MNIST and BinaryNet/CIFAR-10.
            	    Each data point represents a distinct batch size.
            	    BinaryNet/CIFAR-10 training with Keras was not possible due to the Raspberry Pi's memory limit~(\ref{plt:rpi_max_mem}).
                }
            	\label{plot:prototype_mem_trainTime}
            \end{figure*}
            
            \begin{figure*}
                \centering
                \begin{tikzpicture}
    
    \begin{groupplot}[
		ybar,
        scale only axis,
		width=0.35\textwidth,
		height=0.35\textwidth,
		group style={group size=2 by 1, ylabels at=edge left, horizontal sep=5em},
        ylabel near ticks,
        log ticks with fixed point,
        xlabel near ticks,
		xtick={1, 2},
        xticklabels={\makecell{MLP\\MNIST}, \makecell{BinaryNet\\CIFAR-10}},
        enlarge x limits=0.5,
        /pgf/number format/1000 sep={},
        legend image code/.code={
            \draw[#1, bar width=6pt, yshift=-0.3em] plot coordinates {(0cm,0.8em)};
        }
    ]
        
        \nextgroupplot [
    		ylabel=Measured memory (MiB),
            ymin=1,
            ymax=1000,
            ymode=log,
        ]
        
        \node [text width=10em, anchor=north] at (axis description cs:0.5, 1) {\subcaption{Memory\label{plot:prototype_mem}}};
            
        \addplot[thick, pattern=vertical lines, pattern color=black!25!green] coordinates {(1, 258.609375) (2, 1.00)}
        node [pos=0, xshift=-1.2*\pgfkeysvalueof{/pgf/bar width}, rotate=90, anchor=west] {}
        node [pos=1, xshift=-1.2*\pgfkeysvalueof{/pgf/bar width}, rotate=90, anchor=west] {(Std. Keras out of memory)};
        \label{plt:prototype_mem_keras}
        \addplot[thick, pattern=north west lines, pattern color=black!25!cyan] coordinates {(1, 11.25829315) (2, 474.234375)}
        node [pos=0, xshift=0.0*\pgfkeysvalueof{/pgf/bar width}, rotate=90, anchor=west] {22.97$\times$};
        \label{plt:prototype_mem_optstd}
        \addplot[thick, pattern=horizontal lines, pattern color=orange] coordinates {(1, 5.000389099) (2, 154.3671875)}
        node [pos=0, xshift=1.2*\pgfkeysvalueof{/pgf/bar width}, rotate=90, anchor=west] {2.25$\times$}
        node [pos=1, xshift=1.2*\pgfkeysvalueof{/pgf/bar width}, rotate=90, anchor=west] {3.07$\times$};
        \label{plt:prototype_mem_optours}
        
        \nextgroupplot [
    		ylabel=Energy/batch (J),
            ymin=0.1,
            ymax=1000,
            ymode=log,
            log origin=infty,
        ]
        
        \node [text width=10em, anchor=north] at (axis description cs:0.5, 1) {\subcaption{Energy\label{plot:prototype_energy}}};
            
        \addplot[thick, pattern=vertical lines, pattern color=black!25!green] coordinates {(1, 0.676873596) (2, 0.10)}
        node [pos=0, xshift=-1.2*\pgfkeysvalueof{/pgf/bar width}, rotate=90, anchor=west] {}
        node [pos=1, xshift=-1.2*\pgfkeysvalueof{/pgf/bar width}, rotate=90, anchor=west] {(Std. Keras out of memory)};
        \label{plt:prototype_energy_keras}
        \addplot[thick, pattern=north west lines, pattern color=black!25!cyan] coordinates {(1, 1.45125) (2, 160.6909091)}
        node [pos=0, xshift=0.0*\pgfkeysvalueof{/pgf/bar width}, rotate=90, anchor=west] {0.47$\times$};
        \label{plt:prototype_energy_optstd}
        \addplot[thick, pattern=horizontal lines, pattern color=orange] coordinates {(1, 1.417322835) (2, 136.5278527)}
        node [pos=0, xshift=1.2*\pgfkeysvalueof{/pgf/bar width}, rotate=90, anchor=west] {1.02$\times$}
        node [pos=1, xshift=1.2*\pgfkeysvalueof{/pgf/bar width}, rotate=90, anchor=west] {1.18$\times$};
        \label{plt:prototype_energy_optours}
        
        \node[anchor=north,draw,fill=white,inner sep=0.2em] at (axis description cs:0.33, 0.85) {\small
        \begin{tabular}{cl}
            \ref{plt:prototype_mem_keras} & Std. (Keras)\\
            \ref{plt:prototype_mem_optstd} & Std. (CBLAS)\\
            \ref{plt:prototype_mem_optours} & Prop. (CBLAS)
        \end{tabular}};

    \end{groupplot}
    
\end{tikzpicture}
            	\caption{
            	    Measured peak memory consumption \subref{plot:prototype_mem} and energy consumption per batch \subref{plot:prototype_energy} for implementations training MLP/MNIST and BinaryNet/CIFAR-10.
            	    Batch sizes of 200 and 40 were chosen for MLP and BinaryNet, respectively.
            	    BinaryNet/CIFAR-10 training with Keras was not possible due to the Raspberry Pi's memory limit.
            	    Annotations show decreases \emph{vs} the bar to the left.
                        The energy savings in \subref{plot:prototype_energy} were less significant than memory savings in \subref{plot:prototype_mem}, since the memory traffic-associated energy reductions are partially offset by the costs of {\tt bool}-packing (and -unpacking) operations.
                }
            	\label{plot:prototype_mem_and_energy}
            \end{figure*}
            
            Figs.~\ref{plot:prototype_mem_trainTime_mlp} and \ref{plot:prototype_mem_trainTime_binarynet} show the measured memory footprint \emph{vs} training time for the na\"ive (standard and proposed) and Keras implementations across a range of batch sizes.
            For MLP trained to classify MNIST, our na\"ive implementation saw memory requirements reduce by 2.90--4.54$\times$ \emph{vs} the standard approach, with no impact on speed.
            While use of Keras led to much shorter training times, this came at the cost of superproportional memory increases: two orders of magnitude higher than the demands of the proposed approach.
            Keras-based training of BinaryNet is not possible due to the platform's 1~GiB memory limit.
            Keras' training backend uses methods which buffer additional copies of data to optimize for training speed and, as far as we know, the option is not exposed for parametrization~\cite{KERAS_MEMORY_ISSUE}.
            
        
        \subsubsection{CBLAS Acceleration}
        
            In a bid to close our training time gap with Keras, we optimized our prototypes using the CBLAS library, trading off memory for speed~\cite{blas}.
            As shown in Fig.~\ref{plot:prototype_mem_trainTime_mlp}, this reimplementation led to reductions in training times of an order of magnitude with MLP, making our optimized implementations reach similar speed to Keras.
            While the CBLAS-accelerated proposed algorithm requires 1.59--2.08$\times$ more memory than its na\"ive counterpart, this comes in return for speedups of 8.60--29.76$\times$ while remaining 2.16--2.61$\times$ more memory-efficient than the standard approach with acceleration.
            Our approach with CBLAS bettered Keras' memory requirements by 27.66--58.34$\times$ while experiencing slowdowns of 2.10--3.22$\times$.
            Experiments with BinaryNet and CIFAR-10 showed similar trends, with the accelerated standard implementation failing to run with a batch size over 40.
            Note that, due to operating system overheads, it was not possible for the running training program to occupy all of the platform's memory.
            In our CBLAS implementation, the additional data format conversions between floating point and boolean were efficiently accelerated with ARM's single-cycle VCVT instructions. 
            ARM also features native support for fp16 format with VFPv3 architecture in more advanced devices, which would further advance our memory savings.
            
        \paragraph{Energy Efficiency}
        
            In addition to memory savings, our use of low-precision activations and gradients also reduces memory traffic, leading to reduced energy consumption.
            Fig.~\ref{plot:prototype_mem_and_energy} shows the measured memory footprint and energy consumption per epoch for both MLP with MNIST and BinaryNet with CIFAR-10.
            For the batch sizes we tested, the CBLAS-accelerated implementation of our proposed training method surpasses the equally optimized standard approach in terms of energy efficiency by 1.02$\times$ and 1.18$\times$ for those respective network-dataset pairs.
            We remark that these savings shown in Fig.~\ref{plot:prototype_energy} are not as significant when compared against the huge memory reductions shown in Fig.~\ref{plot:prototype_mem}, since data movement cost only accounts for a portion of the overall energy cost, and the memory traffic-associated energy reductions are partially offset by the costs of {\tt bool}-packing (and -unpacking) operations at output (and input) to \emph{every} non-{\tt float32} GEMM kernel.
            Due to lack of an assembly level-optimized bit-packing operation in CBLAS library, we opt to implement in our prototypes a C++-based function which revisits all input and output data to the GEMM kernels, leading to extra data movements.
            This overhead can be reduced by customizing the CBLAS GEMM implementation to perform bit packing (and unpacking) on the fly.

\section{Conclusion}

    In this article, we introduced a neural network training scheme tailored specifically to BNNs.
    Moving first to 16-bit floating-point representation, we selectively and opportunistically approximated beyond this based on careful analysis of the standard training algorithm presented by Courbariaux \& Bengio~\cite{BNN_CNN_BinaryNet}.
    With a comprehensive evaluation conducted across multiple models, datasets, optimizers, and batch sizes, we showed the generality of our approach and reported significant memory reductions \emph{vs} the prior art, challenging the notion that the resource constraints of edge platforms present insurmountable barriers to on-device learning.
    We validated the veracity of our claimed savings with Raspberry Pi-targeted prototypes, whose source code we have made openly available for use and further development.
    In the future, we will explore the potential of our training approximations in the custom hardware domain, within which we expect there to be vast energy-saving opportunity via use of tailor-made arithmetic operators.
    
\section{Acknowledgments}

    The authors are grateful for the support of the United Kingdom EPSRC (grant numbers EP/P010040/1 and EP/S030069/1).
    They also wish to thank Sergey Ioffe and Michele Covell for their helpful suggestions.
    
    For the purpose of open access, the authors will apply a Creative Commons Attribution (CC BY) license to any accepted version of this manuscript.
    

\newpage
\bibliographystyle{ACM-Reference-Format}
\bibliography{bibliography}

\end{document}